\documentclass{article} 
\usepackage{iclr2025_conference,times}
\usepackage{booktabs}   
\usepackage{graphicx}

\usepackage{amsmath,amsfonts,bm}

\def\eqref#1{equation~\ref{#1}}

\def\1{\bm{1}}

\DeclareMathAlphabet{\mathsfit}{\encodingdefault}{\sfdefault}{m}{sl}
\SetMathAlphabet{\mathsfit}{bold}{\encodingdefault}{\sfdefault}{bx}{n}

\usepackage{hyperref}
\usepackage{url}
\usepackage{float}
\usepackage{listings}
\usepackage{fvextra}
\usepackage{subcaption}

\usepackage{tikz}
\usetikzlibrary{arrows, positioning, decorations.pathreplacing, calc}

\title{WILT: A Multi-turn, Memorization-Robust \\ Inductive Logic Benchmark for LLMs}

\author{Eryk Banatt, Jonathan Cheng, Skanda Vaidyanath, Tiffany Hwu \\
Riot Games\\
Los Angeles, CA 90064, USA \\
\texttt{\{ebanatt,joncheng,svaidyanath,thwu\}@riotgames.com} \\
}

\iclrfinalcopy 
\begin{document}

\maketitle

\begin{abstract}
While large language models (LLMs) have shown impressive capabilities across a wide range of domains, they still encounter significant challenges in reasoning tasks that require gathering evidence over multiple turns and drawing logical conclusions from this evidence. These challenges present significant obstacles for LLM chat user interfaces, which rely on multi-turn interactions to facilitate effective collaboration. This limitation leads to real-world issues; for example, service chatbots must gather necessary information from customers over multiple turns to diagnose and resolve problems effectively. Despite the multi-turn nature of many real-world LLM use cases, most existing benchmarks rely on carefully curated single-turn tests, which often blur the line between memorization and genuine reasoning. To address this, we introduce the \textbf{Wason Inductive Logic Test (WILT)}, a simple yet challenging multi-turn reasoning benchmark designed to resist memorization. WILT is inspired by the Wason 2-4-6 task \citep{wason1960failure}, where participants must infer a basic boolean function involving three variables (e.g., $x < y < z$) by proposing test cases (such as $(2, 4, 6)$). In WILT, each test starts from a clean slate, with only the initial instructions provided, preventing models from relying on pre-learned responses. Over several turns, models must interact with the environment by suggesting test cases to narrow the possible hypotheses and ultimately infer the hidden function based on the outcomes. Our findings reveal that LLMs struggle with this task, exhibiting distinct strengths and weaknesses: some are better at narrowing down the hypothesis space by proposing valuable test cases, while others are more adept at deducing the hidden function from observed cases. Despite these variations, the best-performing model achieves only 28\% accuracy, highlighting a significant gap in LLM performance on complex multi-turn reasoning tasks.
\end{abstract}

\section{Introduction}

Large language models (LLMs) powered by the transformer architecture \citep{vaswani2017attention} have enabled a new computing paradigm driven by natural language. These models are increasingly integrated into day-to-day life beyond the machine learning research space, where they help many people with common tasks. These models interact with users through \textit{multi-turn conversations}, a capability of next-token-prediction models bolstered via instruction-tuning \citep{mishra2021cross} and alignment post-training phases \citep{ouyang2022training}. Despite their impressive single-turn performance, LLMs often falter in scenarios requiring multi-turn reasoning and evidence gathering.

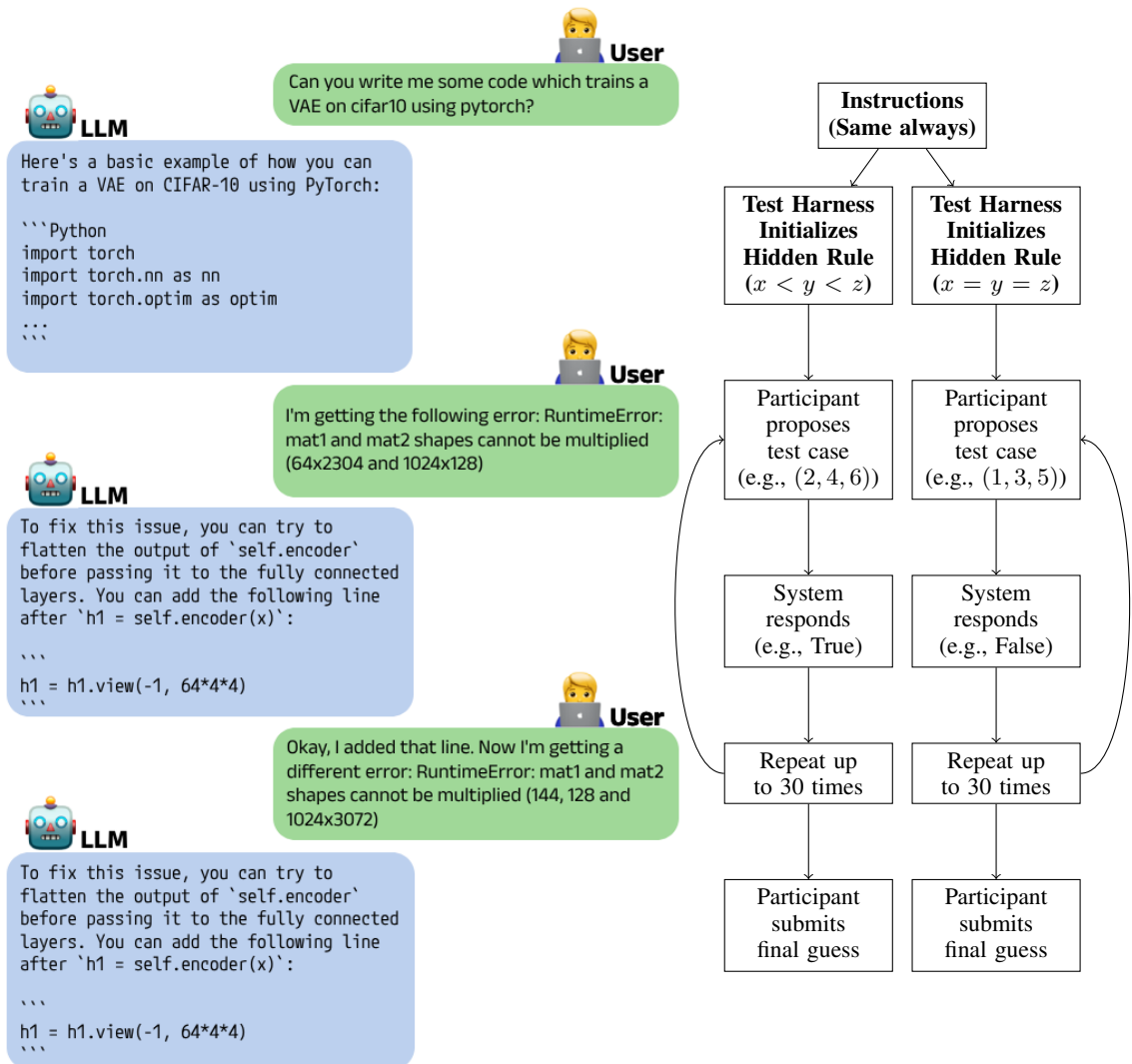
\begin{figure*}[ht]
\hspace{-0.25cm}
\begin{subfigure}[b]{0.65\textwidth}
    \centering
    \includegraphics[width=\textwidth]{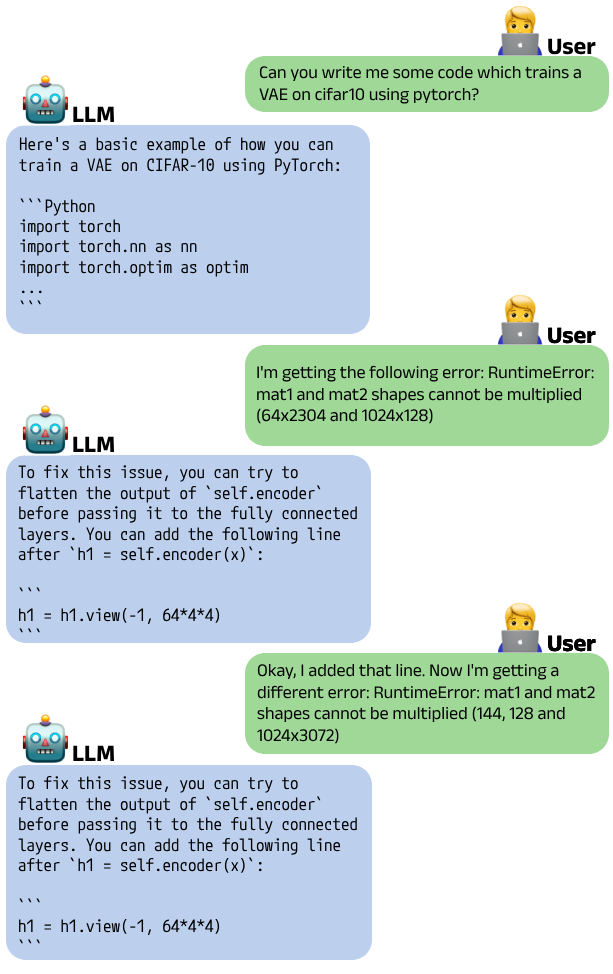}
\end{subfigure}%
\hspace{-0.5cm}
\begin{subfigure}[b]{0.3\textwidth}
    \centering
    \raisebox{1.5cm}{
    \begin{tikzpicture}[node distance=1cm and 1cm, auto, font=\footnotesize]
        \node[rectangle, draw, text width=2cm, align=center] (instructions) {\textbf{Instructions (Same always)}};
    
        \node[rectangle, draw, text width=2cm, align=center, below left=0.5cm and -1cm of instructions] (initialization1) {\textbf{Test Harness Initializes Hidden Rule ($x < y < z$)}};
        \node[rectangle, draw, text width=2cm, align=center, below=1cm of initialization1] (testcase1) {Participant proposes test case \\ (e.g., $(2, 4, 6)$)};
        \node[rectangle, draw, text width=2cm, align=center, below=1cm of testcase1] (response1) {System responds \\ (e.g., True)};
        \node[rectangle, draw, text width=2cm, align=center, below=1cm of response1] (attempts1) {Repeat up to 30 times};
        \node[rectangle, draw, text width=2cm, align=center, below=1cm of attempts1] (finalguess1) {Participant submits final guess};
    
        \node[rectangle, draw, text width=2cm, align=center, below right=0.5cm and -1cm of instructions] (initialization2) {\textbf{Test Harness Initializes Hidden Rule ($x = y = z$)}};
        \node[rectangle, draw, text width=2cm, align=center, below=1cm of initialization2] (testcase2) {Participant proposes test case \\ (e.g., $(1, 3, 5)$)};
        \node[rectangle, draw, text width=2cm, align=center, below=1cm of testcase2] (response2) {System responds \\ (e.g., False)};
        \node[rectangle, draw, text width=2cm, align=center, below=1cm of response2] (attempts2) {Repeat up to 30 times};
        \node[rectangle, draw, text width=2cm, align=center, below=1cm of attempts2] (finalguess2) {Participant submits final guess};
    
    
        \draw[->] (instructions) -- (initialization1);
        \draw[->] (initialization1) -- (testcase1);
        \draw[->] (testcase1) -- (response1);
        \draw[->] (response1) -- (attempts1);
        \draw[->] (attempts1) -- (finalguess1);
        \draw[->] (attempts1.west) to [out=180,in=180,looseness=0.5] (testcase1.west);
    
        \draw[->] (instructions) -- (initialization2);
        \draw[->] (initialization2) -- (testcase2);
        \draw[->] (testcase2) -- (response2);
        \draw[->] (response2) -- (attempts2);
        \draw[->] (attempts2) -- (finalguess2);
        \draw[->] (attempts2.east) to [out=0,in=0,looseness=0.5] (testcase2.east);
    
    \end{tikzpicture}
    }
\end{subfigure}
\vspace{-0.25cm}
\caption{Left: Abbreviated example of the ``doom loop'' phenomenon observed in Appendix \ref{appendix:doom-loop-example}. The model returns an unsatisfactory response on an early turn, and devolves into repeating the same suggestions in subsequent turns. Right: Overview of the WILT task. For each test, the test harness initializes a hidden rule, and participants propose up to 30 test cases for each hidden rule before making a final guess.}
\label{fig:wilt-overview}
\end{figure*}

The reasoning capability of LLMs, particularly in multi-turn scenarios, is of substantial interest. 
A commonly reported failure pattern for LLMs is the ``doom loop'', where after an initially unsatisfactory response, the model repeatedly responds with a near-identical message to one of its earlier messages, providing minimal utility. 
For example, in code generation tasks like the example shown in Figure~\ref{fig:wilt-overview}, a model may repeatedly reply with code blocks identical to ones it has already seen or produced, which impairs its subsequent usefulness. 
Strong multi-turn performance can be thought of as collecting evidence over multiple steps to reduce the hypothesis space of the model's possible responses to only include useful ones: an important capability which remains critical beyond the first turn even in everyday LLM use cases. 

Measuring the performance of LLMs has been challenging for the research community. Existing benchmarks overwhelmingly focus on single-turn interactions \citep{hendrycks2020measuring, hendrycks2021measuring}, and are often subject to memorization or overfitting \citep{goodhart1984problems, recht2019imagenet}. As a result, difficult benchmarks often keep a held-out, publicly unavailable test set to accurately evaluate models \citep{chollet2019measure}. While some multi-turn benchmarks have emerged to attempt to measure phenomena like the ``doom loop'' \citep{mclaughlin2024aidanbench}, none so far have done so for multi-turn, multi-step inductive reasoning.

To bridge this gap, we introduce the \textbf{Wason Inductive Logic Test (WILT)}, a multi-turn benchmark designed to assess LLMs' ability to gather evidence and reason across multiple turns. Inspired by the Wason 2-4-6 task \citep{wason1960failure}, WILT requires models to propose test cases (e.g. $(2,4,6)$), observe the results (True or False), and iteratively narrow down the hypothesis space in order to deduce a hidden rule (e.g. $x < y < z$). 
Since the task always begins from an identical starting state, it cannot simply be solved by memorizing the answer to a prompt and forces the model to use multiple turns to gather evidence.
Therefore, WILT explicitly tests for two important capabilities in an LLM: (1) whether the model can efficiently narrow down the hypothesis space of possible solutions by posing the right test cases over multiple turns, and (2) whether the model can recommend a simple and meaningful solution to the problem after accumulating evidence. While we find that LLMs struggle on this task overall, we find that some of them are better on one of aforementioned sub-tasks than the other. This knowledge can be useful for practitioners when choosing which models to deploy for real world use cases.



In this work, we make the following contributions:

\begin{enumerate}
    \item \textbf{Introducing WILT} - We introduce a multi-turn inductive logic benchmark that cannot be solved in a single turn. This requires a model to gather information, reflect upon what it observes, and make decisions when sufficient evidence is gathered, mirroring real-world tasks like debugging code or reasoning over time. 
    \item \textbf{Evaluating LLMs} - We show that state-of-the-art LLMs struggle significantly on this task, with the best model achieving only 28\% accuracy. This highlights a significant deficiency in multi-turn performance of top-performing LLMs, suggesting that strong single-turn performance may not always transfer to multi-turn settings. 
    \item \textbf{Further Analysis} - We perform several experiments analyzing hypothesis space reduction, response complexity, and the performance of models when provided test cases from a different model. We demonstrate that models have varied strengths and weaknesses on this task, and that combining models together in a multi-turn setting may increase performance compared to using a single strong model. 
\end{enumerate}

\section{WILT}

The Wason Inductive Logic Test (WILT) is a benchmark for LLM reasoning inspired by the Wason 2-4-6 task \citep{wason1960failure}. Models begin with the instruction that they must uncover the hidden rule and may pose up to 30 test cases of that rule. For example, they can pose the tuple $(2,4,6)$ and the test will respond with ``True, 29 Attempts Remaining.'' 

All hidden rules take three numbers and return a boolean. These rules are simple and non-stochastic, so there is no additional value to posing the same test multiple times. Valid inputs include any float or integer that can be typed in three or fewer characters, excluding signs and the decimal point (e.g. -999, 1.23, 5). The hidden rules are written as Python lambda functions. After a maximum of thirty tries (or any turn before then), the model may make one attempt to guess the function, after which the test will terminate. The model must return a Python lambda function that is the same as or equivalent\footnote{For example, $x - y = z$ is equivalent to $y + z = x$} to the hidden rule in order to receive full points.

WILT is conceptually simple, but very challenging. Humans are able to identify simple rules despite the infinitely large hypothesis space, the unbounded difficulty of a hidden function, and the impossibility of verifying the correctness of your response \citep{tweney1980strategies, tukey1986philosophical}.
Without effective priors for what constitutes a ``simple'' or ``reasonable'' answer, it becomes trivial to construct an arbitrary counterexample to a current hypothesis. 
For example, consider the canonical Wason task rule of $x < y < z$. This rule has very high overlap with the much more arbitrary rule $(x < y < z) \land (x \neq 12)$. 

The WILT benchmark therefore tests a few high-value behaviors of interest:

\begin{enumerate}
    \item \textbf{Multi-Turn Capability}: Participants that fall into doom loops are punished by virtue of having less useful information with which to infer the hidden rule.
    \item \textbf{Hypothesis Space Reduction}: Participants are rewarded for proposing test cases that effectively narrow down the possible rules, despite that hypothesis space being infinitely large.
    \item \textbf{Susceptibility to Confirmation Bias}: Participants who are more prone to ``confirming their hypothesis'' rather than seeking falsification will perform poorly upon this task.
    \item \textbf{Inductive Reasoning}: Proposing good test cases is a useful test of inductive reasoning and the ability to generalize from a number of specific examples. 
    \item \textbf{Deductive Reasoning}: Proposing sensible functions after observing many test cases is a useful test of deductive reasoning.  Success in this task involves identifying a specific function that accurately fits a set of examples.
    \item \textbf{Occam's Razor}: Participants are rewarded for finding the simplest explanation fitting the examples.
\end{enumerate}

We release two test suites: a \textit{lite split}, with 10 very easy tests and a canonical \textit{full split} with 50 moderately difficult tests\footnote{We will release the code for this benchmark with the final release of the full paper.}. Future work will extend this to include a procedurally generated split for additional robustness to overfitting. We find that the lite split quickly produces a roughly similar ordering to the full split, but we report results upon the full split for the remainder of this work. Please see Appendix \ref{appendix:lite-split} for further details.


\section{Related Work}

Compared to other reasoning benchmarks, WILT stands out as both highly multi-turn focused and unusually robust to memorization. In contrast to other benchmarks, WILT requires models to \textit{interact with an environment} by proposing their own test cases to uncover a hidden function without relying on pre-provided examples. This setup reduces the risk of overfitting, as each test begins with the same initial instructions, and the model must generate and interpret its own data. 

\subsection{Reasoning Benchmarks}

There are a wide variety of reasoning benchmarks used to evaluate large language models. Some very notable among these are MATH \citep{hendrycks2021measuring}, GSM8K \citep{cobbe2021training}, CommonsenseQA \citep{talmor2018commonsenseqa}, StrategyQA \citep{geva2021did}, BIG-BENCH \citep{srivastava2022beyond}, SciBench \citep{wang2023scibench}, SVAMP \citep{patel2021nlp}, ARC-AGI \citep{chollet2019measure}, MMLU \citep{hendrycks2020measuring}, GPQA \citep{rein2023gpqa}, and HumanEval \citep{chen2021evaluating}. These benchmarks are the standard for measuring LLM reasoning capabilities, but are overwhelmingly carefully chosen single-turn problems which aim to meaningfully separate the performance of different models on reasoning-like outputs such as math, code, or logic puzzles. However, these benchmarks are subject to train-on-test leakage, even if efforts are made to decontaminate the dataset \citep{yang2023rethinking}, and the majority are explicitly single-turn tests. Our benchmark directly measures the model's ability to navigate multi-turn scenarios and does not require careful hiding of a test set to prevent misleading results. 

With respect to reasoning about simple functions, a benchmark that stands out as similar to ours is CRUXEval \citep{gu2024cruxeval}, which assembles a list of 800 simple Python functions and input-output pairs, evaluating language models on their ability to predict input from output and output from input. Our work could be seen as a multi-turn, more difficult extension of this work -- one where the function is replaced with a black box, where helpful and informative input-output pairs are not provided but instead need to be searched for by the language model, and where the objective is to infer the hidden function rather than the input or output.

\subsection{Multi-Turn Benchmarks}

There are a handful of multi-turn benchmarks used to evaluate LLMs. PlanBench \citep{valmeekam2022large} is one prominent benchmark that attempts to measure the ability of LLMs to navigate planning problems. This is a class of problems that is solved easily by classical planning algorithms such as STRIPS \citep{fikes1971strips}, and like our benchmark poses a significant challenge to LLMs. PlanBench is a primarily multi-step, single-turn benchmark with a multi-turn component (i.e. replanning based on unexpected events), which contrasts with our benchmark's more direct multi-turn focus. This can be observed in the o1 models performing comparatively well on PlanBench \citep{valmeekam2024llmscantplanlrms}, since scaling inference time compute within a single turn would be expected to improve performance substantially.

Closest to ours are MINT \citep{wang2023mint} and Aidan-bench \citep{mclaughlin2024aidanbench}, which have more direct multi-turn focus. MINT repurposes existing single-turn benchmarks by allowing models to use tools before answering. While the range of tasks in MINT is therefore quite large, strong models can still solve these tasks in few (or one) turns, and the unmodified prompts remain subject to test set leakage. Aidan-bench measures the cosine similarity between multi-turn responses. This represents a more pure measurement of the doom loop phenomenon. In our benchmark, rather than directly measuring the doom loops, we are instead measuring how often those doom loops lead to failures of reasoning. We see similar surprising performances in our benchmark compared to Aidan-bench (e.g. Mistral Large), but with an ordering more tied to capabilities (e.g. Sonnet's strong results, see Table \ref{tab:model-accuracy}).

\subsection{Hypothesis Space Reduction}

Hypothesis space representation is a commonly used framing in inductive logic tasks for LLMs. In \cite{wang2023hypothesis}, the authors show a technique called \textit{hypothesis search} where the model will propose hypotheses in natural language and then implement these hypotheses as Python programs. This technique was shown to improve performance on ARC-AGI \citep{chollet2019measure}, but a similar approach could be used along with chain-of-thought \citep{wei2022chain} for WILT as well.

\section{Experiments}

\subsection{WILT Full split Results}

We evaluated several state-of-the-art LLMs on the full split of the WILT task. Experimental details can be found in Appendix \ref{appendix:experimental-details}. Our results for this test can be found in Table \ref{tab:model-accuracy}. Claude 3.5 Sonnet narrowly performs the best on this task, with OpenAI's o1 models close behind despite using fewer guesses.   

\begin{table}
  \caption{Model Accuracy Comparison}
  \label{tab:model-accuracy}
  \centering
  \begin{tabular}{lllll}
    \toprule
    \textbf{Model} & \textbf{Accuracy}  & \textbf{Approx.} & \textbf{Avg.} & \textbf{Avg.}\\
    & & \textbf{Correct} & \textbf{Guesses} & \textbf{Repeats}\\
    \cmidrule(r){1-5}
        Claude 3.5 Sonnet \citep{claude3.5sonnet2024} & \textbf{14/50} & \textbf{10/50} & 16.38 & 0.54 \\
        o1-mini 2024-09-12 \citep{openai2024o1system} & 13/50 & 8/50 & 12.1 & \textbf{0.06} \\
        o1-preview 2024-09-12 \citep{openai2024o1system} & 12/50 & 6/50 & \textbf{8.12}\footnotemark & \textbf{0.06}\\
        chatgpt-4o-latest \citep{gpt4o2024} & 11/50 & 7/50 & 14.22 & 0.76 \\
        Mistral Large 2 \citep{mistral2407} & 11/50 & 5/50 & 26.56 & 2.84\\
        GPT-4o 2024-08-06 \citep{gpt4o2024} & 9/50 & 6/50 & 15.26 & 0.52\\
        Llama 3.1 405B \citep{dubey2024llama} & 8/50 & 9/50 & 12.21 & 0.60\\
        Gemini 1.5 Flash 0827 \citep{google2024gemini} & 7/50 & 4/50 & 14.04 & 2.16\\
        Llama 3.1 70B \citep{dubey2024llama} & 7/50 & 2/50 & 15.18 & 1.48\\
        Deepseek-v2.5-chat \citep{liu2024deepseek} & 6/50 & 5/50 & 27.22 & 9.78\\
        GPT-4o-mini \citep{gpt4omini2024} & 6/50 & 2/50 & 20.36 & 1.08 \\
        Gemini 1.5 Pro \citep{google2024gemini} & 5/50 & 6/50 & 16.78 & 0.82\\
        Gemini 1.5 Flash \citep{google2024gemini} & 5/50 & 6/50 & 16.5 & 2.46\\
        Deepseek-v2-coder \citep{zhu2024deepseek} & 5/50 & 5/50 & 21.82 & 6.70\\
        Deepseek-v2-chat \citep{liu2024deepseek} & 3/50 & 3/50 & 25.32 & 6.68\\
        Llama 3.1 8b \citep{dubey2024llama} & 3/50 & 0/50 & 26.18 & 4.46\\
        Open Mistral Nemo \citep{mistralnemo2024} & 2/50 & 3/50 & 27.34 & 8.00\\
        Claude 3 Haiku \citep{claude3haiku2024} & 1/50 & 1/50 & 6.76 & 0.22\\
        Gemini 1.5 Flash 8b 0827 \citep{google2024gemini} & 0/50 & 2/50 & 26.76 & 7.72\\
        Gemma 2 9B \citep{team2024gemma} & 0/50 & 2/50 & 8.82 & 1.40\\
        \bottomrule
  \end{tabular}
\end{table}
\footnotetext{We bold results for this column only if the models are also high performing. For example, Claude 3 Haiku uses fewer guesses than o1-preview, but this is because it is failing to make use of its available turns to uncover the behavior of the hidden rule, not because it efficiently solves the task.}

In Table \ref{tab:model-accuracy}, we include a column \textit{approximately correct}, measuring the number of rules in which the model was able to correctly identify some critical behavior of the rule, but returned a rule with failing edge cases\footnotemark[3]. For example, guessing $(x < y < z)$ instead of $(x \leq y \leq z)$ is approximately correct. We include this column to highlight models that are more willing to guess immediately instead of uncovering edge cases by default (e.g. Llama 3.1 405B). In these cases, we could see potentially improved performance through explicitly instructing the model to use more guesses. Examples of common failure patterns can be found in Appendix \ref{appendix:failures}.

In addition, we find that LLMs (particularly smaller ones) will frequently repeat tests they have already used, sometimes dozens of times, often with the same provided reasoning. We therefore also provide a column \textit{average repeats}, which counts the average number of proposed tests already tested for that rule. Further discussion on test novelty can be found in Appendix \ref{appendix:novelty}.

In our experiments, we show that LLMs struggle substantially with this task. 
We show that despite the test's relative simplicity, most models struggle substantially both to propose good tests and to infer a rule based on available evidence.
Specifically, their propensity to repeat test cases, propose useless test cases, and guess very unlikely rules harms their performance on this task substantially. The varying performance in a multi-turn setting represents a previously unappreciated dimension of measuring reasoning capability in LLMs. There is much work in language modeling for code-based agents \citep{cognitiondevin2024, yang2024swe} and LLM-driven unit testing \citep{yang2024empirical}, and the difficulty of LLMs to explore edge cases effectively across multiple turns has substantial implications on those applications. 

As in the original Wason 2-4-6 task, we find a common failure mode to be confirmation bias -- a participant will identify a plausible hypothesis and continue to propose tests that attempt to \textit{confirm} it (see Appendix \ref{conf-bias-example}). Stronger models will more explicitly attempt to \textit{falsify} these hypotheses instead. For example, a model that observes the rule is always true when the values are all equal should attempt to test a rule like $x \leq y \leq z$, rather than submitting more cases where the three values are equal.

We show some varying behavior among top performing models, which highlights the multifaceted ways in which a model can succeed on a multi-turn task. Claude 3.5 Sonnet performs quite well on WILT, outperforming models like o1-mini and o1-preview, which tend to win on adjacent single-turn benchmarks. On most turns, WILT is a setting that removes many of the advantages of a model that has scaled up inference-time compute; each turn requires only a moderate amount of consideration, compared to other benchmarks that are \textit{multi-step, single turn} and thereby reward the model substantially for long and elaborate chains of thought. o1-mini and o1-preview show astonishingly low test case reuse, and much lower than average guess count compared to Sonnet 3.5 despite comparable performance, pointing to their strong deductive reasoning skills and recollection of the previous steps in a conversation. However, they ultimately underperform due to gathering less evidence compared to Sonnet 3.5, which is a \textit{multi-turn capabilities failure} despite \textit{not} being a failure on any turn in particular. Put another way, Sonnet 3.5 seems better at understanding that it will be able to leverage future tests, and that its multi-turn performance shines ``between'' turns rather than during them.

In this vein, Mistral Large stands out as having surprisingly strong performance in multi-turn, despite its novelty metrics being below average. That is, it tends to respond with many operators (see Table \ref{tab:complexity}), repeat many more tests than other similar-performing models, and responds with messages with worse novelty metrics as reported in Table \ref{tab:model-novelty} and Figure \ref{fig:novelty-turn}. However, it tends to use most of its available resources, accrue evidence with new tests deep into the conversation, and ultimately deduce the rule very well for a model of its size. The strong result observed from Mistral Large points towards a \textit{multi-turn capabilities success} despite numerous \textit{single-turn failures}, and reveals a previously underappreciated ability of this model. 

\subsection{Hypothesis Space Reduction}

\begin{figure} [ht]
    \centering
    \includegraphics[width=0.7\linewidth]{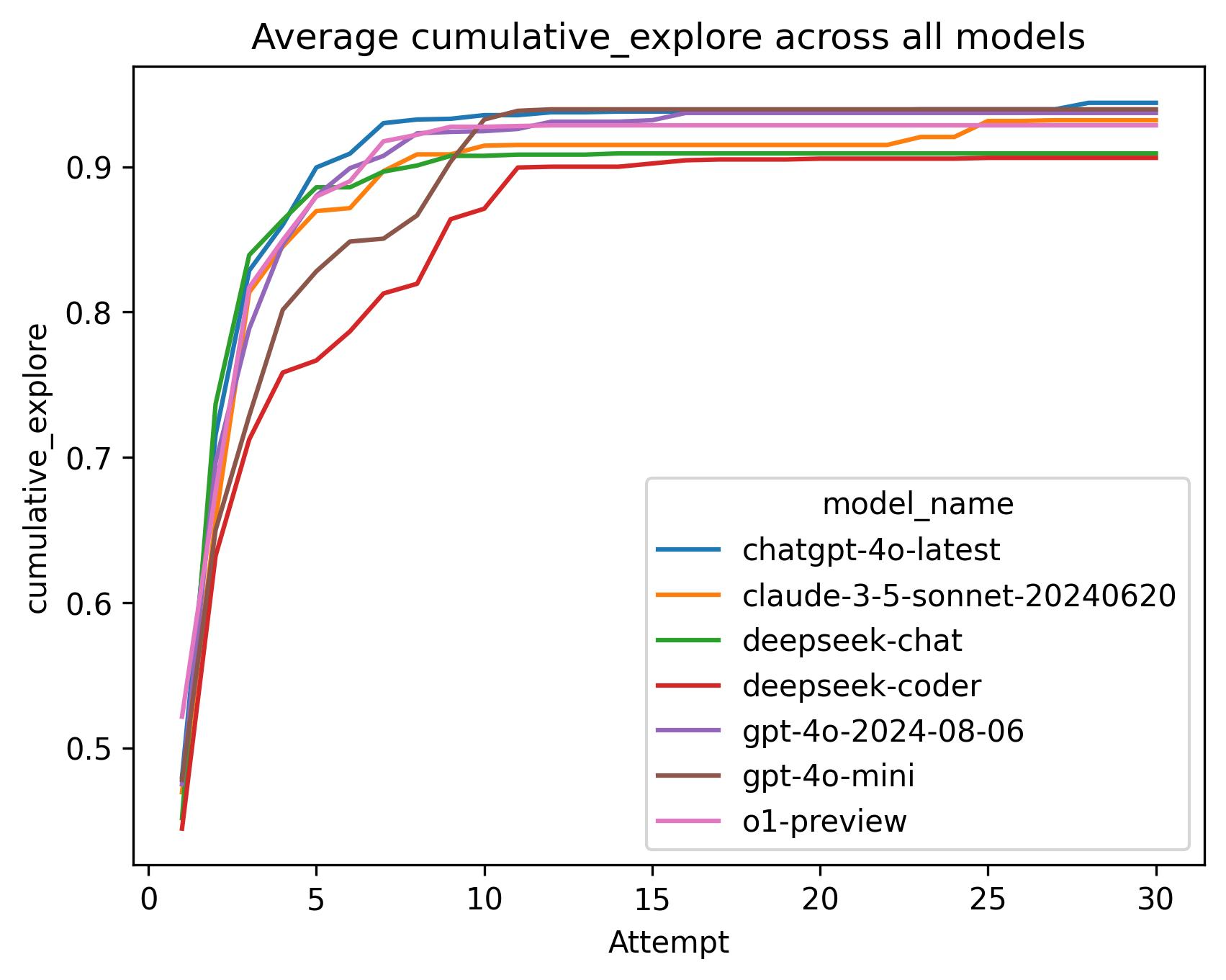}
    \caption{Models can succeed upon this task by reducing the hypothesis space quickly or providing useful tests for many turns. We show that models with strong reasoning capabilities can narrow the space quickly, but weaker multi-turn capability harms their ability to get value out of later tests.}
    \label{fig:hypothesis}
\end{figure}

To compare the LLMs' ability to efficiently reason about the task, we estimate how effectively each model reduces the hypothesis space \citep{wang2023hypothesis}. At best, an LLM should always propose a triplet that eliminates as many untested hypotheses as possible. At worst, a model repeatedly proposes a triplet confirming previously covered hypotheses. For example, an LLM that has already guessed $(2, 4, 6)$ retreads much of the same hypothesis space by guessing $(4, 6, 8)$ rather than $(0.01, -1, 100)$.

To represent that hypothesis space, we randomly generate 2000 lambda functions that encompass a wide range of potential hypotheses. For example, we randomly generate lambdas having to do with: ordering $(x < y < z)$, equality $(x = y \neq z)$, arithmetic relations $(x + y = z)$, parity $(x \leq 10, y \leq 5, z \leq 100)$, etc. When an LLM proposes a triplet, we cross off any lambdas that do not match the observed behavior of the hidden rule. Figure \ref{fig:hypothesis} illustrates the rate at which different models reduce the hypothesis space over successive turns. Models with worse reasoning spend more attempts to clear less of the hypothesis space, with some models using all of their attempts and yet failing to propose useful tests once some visible behavior is latched onto. For a model with strong reasoning capabilities, we would expect that they would not repeatedly propose test cases that cover previously matched lambdas. 
Models like deepseek-v2-chat show strong hypothesis space reduction within the first few turns, but struggle to cover the entire space even after 30 turns, leading to suboptimal performance on the full split. In contrast, models like gpt-4o-mini show weaker performance in the first few turns, but are able to cover a greater range of the hypothesis space by better leveraging later turns. It is important to remember that narrowing the hypothesis space is only one aspect of the challenge; making an accurate final guess after proposing tests is equally crucial for strong overall performance on WILT.

\subsection{Evaluating Function Inversion Capability} \label{split-experiment}

\begin{figure} [ht]
    \centering
    \includegraphics[width=0.8\linewidth]{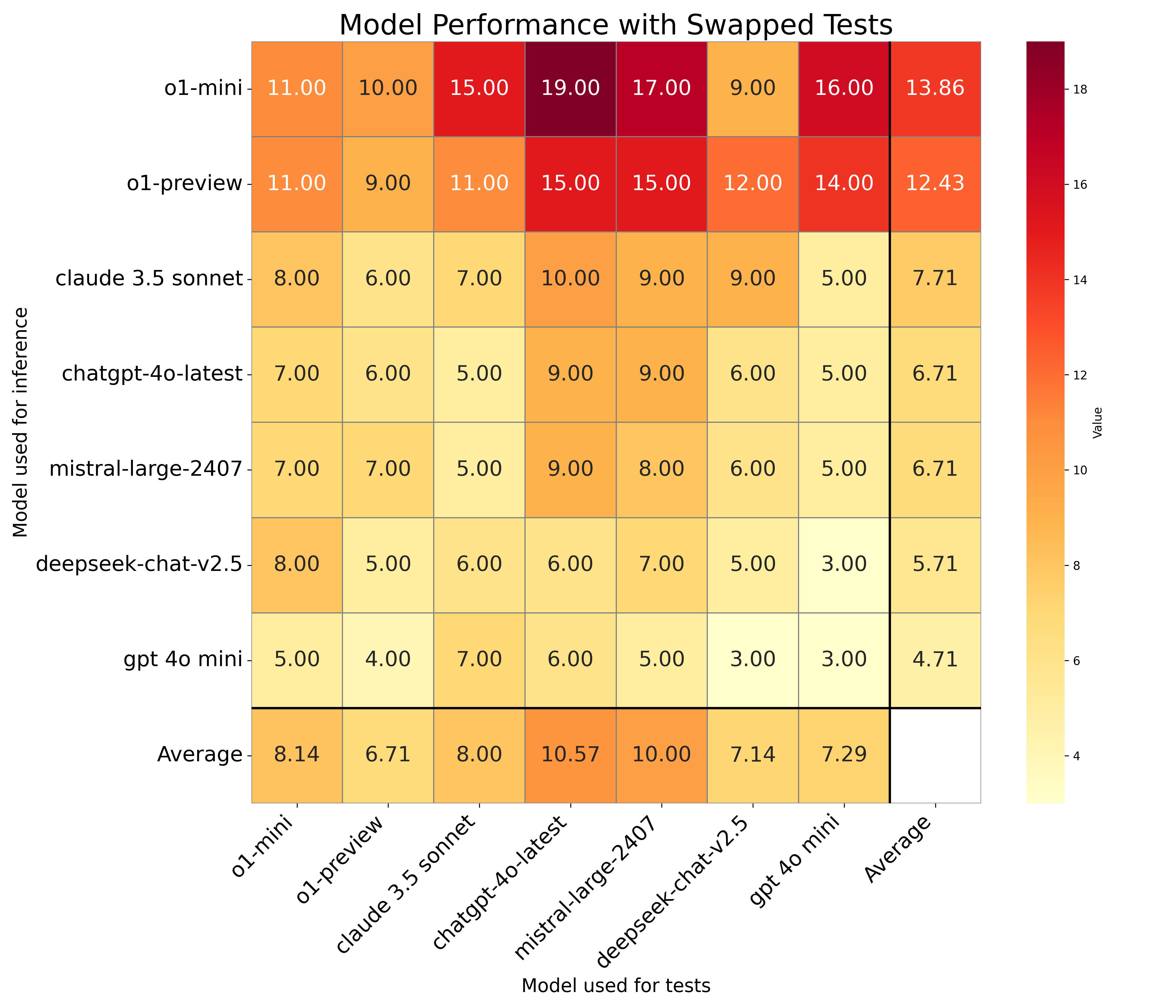}
    \caption{Models have varying success when using test cases proposed by other models. o1-mini stands out as having much stronger single-turn reasoning in this experiment, but it performs poorly with its own tests.
    }
    \label{fig:cross_heatmap}
\end{figure}

To succeed at the WILT task, models must succeed at both gathering evidence (hypothesis reduction) and drawing logical conclusions from evidence (function inversion). 
To distinguish a model's ability to do one or the other, we perform an experiment where models attempt to guess the rule using tests from another model. Rather than asking the model to gather evidence, we directly provide it all the reasoning-stripped\footnote{We strip reasoning to avoid conflating confirmation bias in the attached reasoning, rather than just the accumulated evidence.} input-output pairs generated by another model for the same rule, and ask the model to guess the rule in a single turn. Without the original reasoning and subsequent observation before and after each test case, we expect most models to underperform relative to the full test even when provided their own cases. Likewise, we expect models stronger at single-turn to perform better in this experiment relative to other models subject to the same evidence. Our results can be found in Figure \ref{fig:cross_heatmap}.

This reveals some notably varied capabilities among the top performing models. While Claude Sonnet 3.5 was the narrowly highest performing model on the full test, this experiment reveals important context for why that may be. We see that it performs better than most other models subject to the same evidence, but proposes test cases that are generally slightly less useful for other models without the attached justifications. Likewise, without its own reasoning for each case, Sonnet's performance degraded substantially more than other models in the same setting, suggesting a larger component of its success was its reasoning, compared to the test cases alone.

o1-mini shows highly superior single-turn capability in this test, but notably performs relatively less well when provided its own tests rather than the tests of another high-performing model. When paired with cases from chatgpt-4o-latest, it successfully guessed 19 of the 50 rules, far surpassing the best-performing single model in the full test. 

Despite having many repeated tests and messages which were generally similar to each other (see Tables \ref{tab:model-accuracy} and \ref{tab:model-novelty}), we see that Mistral Large performs well with other models' tests and provides a corpus of tests useful to other models. We note its comparable performance to chatgpt-4o-latest both along the rows (model's performance with other model's test cases) and columns (performance of other models with the model's test cases), reinforcing its strong performance in the full test.

Critically, we show that models have non-identical strengths and weaknesses, and that success on the full WILT task depends on strong performance on a few key metrics of interest. Even without the attached reasoning for test cases, composing the test case generation of one model and the function inversion of another model very often outperforms using a single strong model for both subtasks.
This has some notable implications for future LLM applications: in \cite{wang2024mixture} it was shown that several language models coordinated by an aggregator LLM could outperform strong single models. Future work could explore coordinating models for both single-turn and multi-turn oriented tasks, potentially leading to improved performance.

\subsection{Response Complexity}
To determine how well the models employ Occam's Razor, we explore different metrics to gauge whether the models find the simplest rule that covers the examples. From existing Bayesian models of cognition \citep{tenenbaum1999rules, tenenbaum2001generalization}, the \emph{size principle} uses hypothesis size as a measure of simplicity. In these Bayesian models, hypothesis size is calculated as the number of values that match the hypothesis. Calculating hypothesis size in this manner is only possible when the test values are within a limited countable range. In our case, the possible test values are infinite, requiring some alternative metrics to gauge hypothesis size. We use three metrics: 

\begin{enumerate}
    \item \textbf{Number of Operators}: We count the number of operators used in the rule expression.
    \item \textbf{Response Length}: We calculate the string length of the rule expression. The longer the length, the more complex it is likely to be. As longer outputs tend to be arbitrarily preferred by automatic evaluators \citep{dubois2024length}, it is particularly important to measure the brevity of the response for cases where simplicity is desired.
    \item \textbf{Set Inclusion}: We generate a grid of integer-float tuples and apply them to guessed and actual rules to generate sets of tuples returning ``True''. If the set of the guessed rule is a subset or superset of the actual rule, we then calculate their set size ratio. A ratio of $1$ is ideal, $>1$ suggests a less complex guess, and $<1$ a more complex one. 
\end{enumerate}

Table \ref{tab:complexity} shows the complexity metrics of the LLMs. Many LLMs with high accuracy such as Claude 3.5 Sonnet, chatgpt-4o-latest, and Mistral Large have long response lengths. However, o1-preview has a short response length and few operators, despite its high performance on the task. The differences in response length and number of operators is most clearly seen in the incorrect answers. For example, if the correct rule is \lstinline|lambda x, y, z: (x * y * z) % 2 == 1|, Claude 3.5 Sonnet's guess is \lstinline|lambda x,y,z: all(n > 0 and int(n) == n and (n & (n-1) == 0) and (n % 3 == 0 or n == 1) for n in [x,y,z])|, which is more convoluted than o1-preview's guess of \lstinline|lambda x, y, z: abs(x) == 1 and abs(y) == 1 and abs(z) == 1|. Combined with its low number of average guesses made before making the final guess, o1-preview appears to follow Occam's Razor very well compared to most of the other high-performing models with longer response lengths. For the set inclusion ratio, the best models tend to cluster around an intermediate value of 2-4.


\begin{table}[H]
  \caption{Response Complexity (Median)}
  \label{tab:complexity}
  \centering
  \begin{tabular}{llll}
    \toprule
    \textbf{Model} & \textbf{Num Operators} & \textbf{Response Length} & \textbf{Set Inclusion} \\
    \cmidrule(r){1-4}
    Claude 3.5 Sonnet & 3 & 34.5 & 0.08 \\
    o1-mini 2024-09-12 & 3 & 29.0 & 0.79 \\
    o1-preview-2024-09-12 & 2 & 25 & 0.01 \\
    chatgpt-4o-latest & 5 & 39 & 1.0 \\
    Mistral Large 2 & 5 & 39 & 1.0 \\
    GPT-4o 2024-08-06 & 4.5 & 39 & 0.34 \\
    Llama 3.1 405B & 2 & 30 & 0.52 \\
    Gemini 1.5 Flash 0827 & 4 & 35.5 & 0.00 \\
    Llama 3.1 70B & 2 & 25 & 1.00 \\
    Deepseek-v2.5-chat & 3 & 29 & 0.27 \\
    GPT-4o-mini & 5 & 39.5 & 0.05 \\
    Gemini 1.5 Pro & 3 & 38 & 0.27 \\
    Gemini 1.5 Flash & 3 & 28 & 0.06 \\
    Deepseek-v2-coder & 5 & 39 & 0.88 \\
    Deepseek-v2-chat & 2 & 28 & 0.00 \\
    Llama 3.1 8b & 2 & 23 & 0.05 \\
    Open Mistral Nemo & 5 & 46 & 1.00 \\
    \bottomrule
  \end{tabular}
\end{table}

\begin{figure} [ht]
    \centering
    \includegraphics[width=0.8\linewidth]{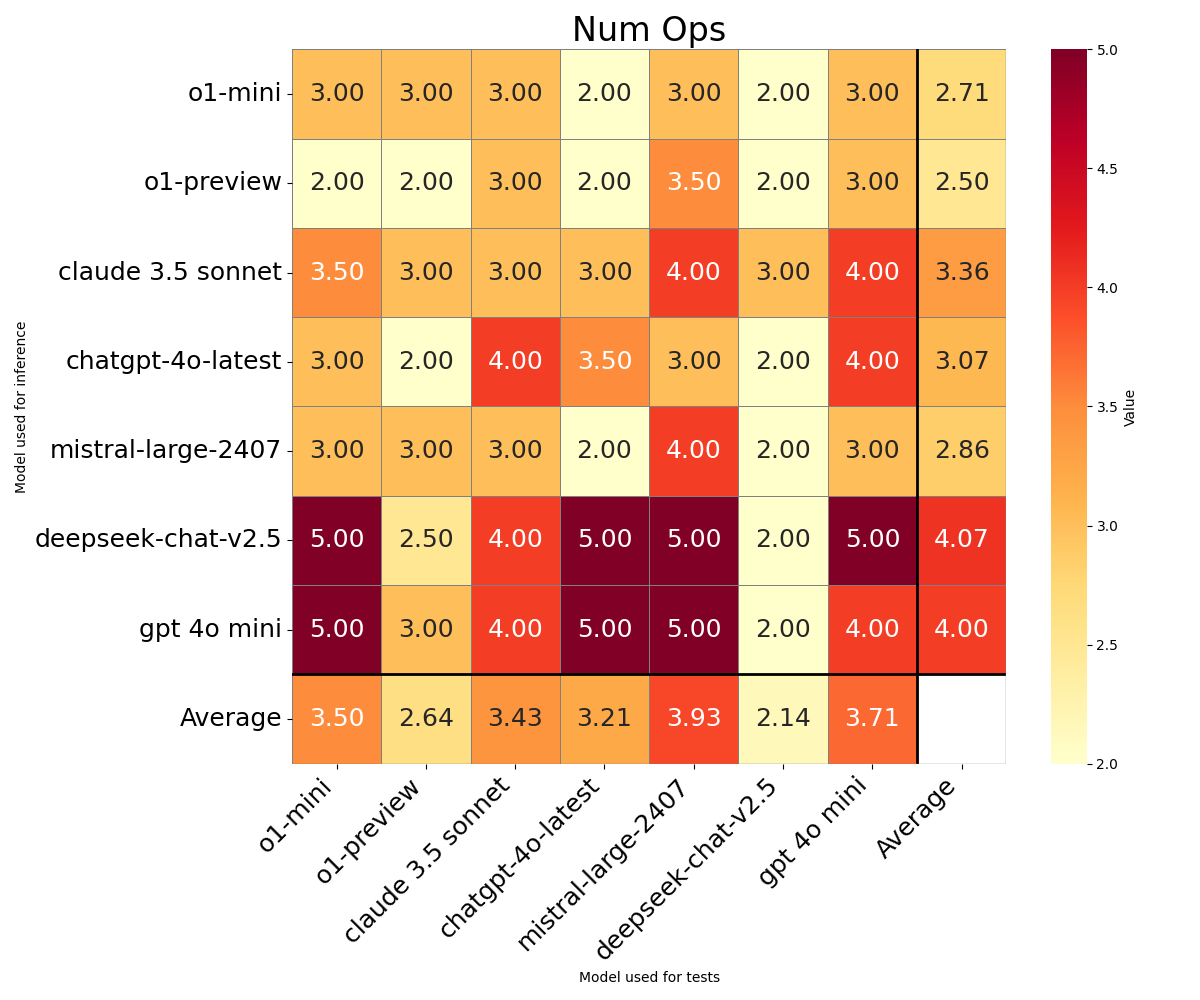}
    \caption{The median number of operators provides a window into guessed rules being more or less complicated compared to guesses provided by other models. More metrics for response complexity in test-case swap scenarios can be found in Appendix \ref{appendix:testswapcomplex}.}
    \label{fig:numops_swap}
\end{figure}

In addition, Figure \ref{fig:numops_swap} shows the median number of operators used by a model when given another model's test cases, as in Section \ref{split-experiment}. This can be used to estimate the model's bias toward guessing a simpler rule. o1-preview, for example, tends to use fewer operators than the other models subject to the same evidence. This also highlights potential discrepancies in the complexity resulting from a model's test cases and the complexity of its final guesses. When models use test cases generated by DeepSeek Chat v2.5, they tend to use fewer operators, likely because the test cases are fully encompassed by simple rules like \verb|lambda x,y,z: False|. Conversely, when given other models' tests, DeepSeek Chat v2.5 responds with a high level of complexity compared to other models. Its guesses often overfit a complicated rule to the test cases (e.g., it guesses a rule of \verb|lambda x, y, z: y == x or y == z or y == (x + z) / 2| when given the following true test cases for even numbers: \lstinline|(2.0, 2.0, 2.0), (2.0, 2.0, 4.0), (-2.0, 4.0, 6.0), (0.0, 2.0, 4.0), (2.2, 4.0, 6.0), (2.0, 4.0, 6.0)|.)

\section{Conclusion}

In this work, we highlight the importance of the multi-turn setting for understanding the performance of large language models in everyday use cases. We introduce a new benchmark that attempts to measure performance in this common setting and show its difficulty even for state-of-the-art LLMs. With the models exhibiting non-overlapping strengths, the resulting performance of these models on the WILT task provides useful context on the types of tasks for which they would be useful, beyond simply suggesting that the top-performing model is the best one to use for all applications. 

Our benchmark identifies a critical way that typical benchmarks differ from day-to-day use of LLMs: whereas effective multi-turn collaboration often requires asking clarifying questions, performing experiments, and gathering new evidence, typical benchmarks are overwhelmingly designed to be solved immediately. This may, in turn, create an implicit incentive for models to end conversations as quickly as possible, even where that may be ineffective or inappropriate. 

With this work, we aim to provide a benchmark that measures a model's capacity for exploring an environment and reasoning based on its own decisions across multiple turns. We believe that this paradigm offers a more direct measurement of reasoning capability compared to other benchmarks. Likewise, we believe an explicit focus on multi-turn reasoning capability more directly shines a light upon the most common uses of LLMs.



\bibliography{iclr2025_conference}

\begin{thebibliography}{48}
\providecommand{\natexlab}[1]{#1}
\providecommand{\url}[1]{\texttt{#1}}
\expandafter\ifx\csname urlstyle\endcsname\relax
  \providecommand{\doi}[1]{doi: #1}\else
  \providecommand{\doi}{doi: \begingroup \urlstyle{rm}\Url}\fi

\bibitem[Abts et~al.(2022)Abts, Kimmell, Ling, Kim, Boyd, Bitar, Parmar, Ahmed,
  DiCecco, Han, et~al.]{abts2022software}
Dennis Abts, Garrin Kimmell, Andrew Ling, John Kim, Matt Boyd, Andrew Bitar,
  Sahil Parmar, Ibrahim Ahmed, Roberto DiCecco, David Han, et~al.
\newblock A software-defined tensor streaming multiprocessor for large-scale
  machine learning.
\newblock In \emph{Proceedings of the 49th Annual International Symposium on
  Computer Architecture}, pp.\  567--580, 2022.

\bibitem[{Anthropic}(2024{\natexlab{a}})]{claude3.5sonnet2024}
{Anthropic}.
\newblock Introducing claude 3.5 sonnet, 2024{\natexlab{a}}.
\newblock URL \url{https://www.anthropic.com/news/claude-3-5-sonnet}.
\newblock Accessed: 2024-09-11.

\bibitem[{Anthropic}(2024{\natexlab{b}})]{claude3haiku2024}
{Anthropic}.
\newblock Claude 3 haiku: Our fastest model yet, 2024{\natexlab{b}}.
\newblock URL \url{https://www.anthropic.com/news/claude-3-haiku}.
\newblock Accessed: 2024-09-11.

\bibitem[Chen et~al.(2021)Chen, Tworek, Jun, Yuan, Pinto, Kaplan, Edwards,
  Burda, Joseph, Brockman, et~al.]{chen2021evaluating}
Mark Chen, Jerry Tworek, Heewoo Jun, Qiming Yuan, Henrique Ponde De~Oliveira
  Pinto, Jared Kaplan, Harri Edwards, Yuri Burda, Nicholas Joseph, Greg
  Brockman, et~al.
\newblock Evaluating large language models trained on code.
\newblock \emph{arXiv preprint arXiv:2107.03374}, 2021.

\bibitem[Chollet(2019)]{chollet2019measure}
Fran{\c{c}}ois Chollet.
\newblock On the measure of intelligence.
\newblock \emph{arXiv preprint arXiv:1911.01547}, 2019.

\bibitem[Cobbe et~al.(2021)Cobbe, Kosaraju, Bavarian, Chen, Jun, Kaiser,
  Plappert, Tworek, Hilton, Nakano, et~al.]{cobbe2021training}
Karl Cobbe, Vineet Kosaraju, Mohammad Bavarian, Mark Chen, Heewoo Jun, Lukasz
  Kaiser, Matthias Plappert, Jerry Tworek, Jacob Hilton, Reiichiro Nakano,
  et~al.
\newblock Training verifiers to solve math word problems.
\newblock \emph{arXiv preprint arXiv:2110.14168}, 2021.

\bibitem[{Cognition AI}(2024)]{cognitiondevin2024}
{Cognition AI}.
\newblock Introducing devin: The first ai software engineer, 2024.
\newblock URL \url{https://www.cognition.ai/blog/introducing-devin}.
\newblock Accessed: 2024-09-11.

\bibitem[Dubey et~al.(2024)Dubey, Jauhri, Pandey, Kadian, Al-Dahle, Letman,
  Mathur, Schelten, Yang, Fan, et~al.]{dubey2024llama}
Abhimanyu Dubey, Abhinav Jauhri, Abhinav Pandey, Abhishek Kadian, Ahmad
  Al-Dahle, Aiesha Letman, Akhil Mathur, Alan Schelten, Amy Yang, Angela Fan,
  et~al.
\newblock The llama 3 herd of models.
\newblock \emph{arXiv preprint arXiv:2407.21783}, 2024.

\bibitem[Dubois et~al.(2024)Dubois, Galambosi, Liang, and
  Hashimoto]{dubois2024length}
Yann Dubois, Bal{\'a}zs Galambosi, Percy Liang, and Tatsunori~B Hashimoto.
\newblock Length-controlled alpacaeval: A simple way to debias automatic
  evaluators.
\newblock \emph{arXiv preprint arXiv:2404.04475}, 2024.

\bibitem[Fikes \& Nilsson(1971)Fikes and Nilsson]{fikes1971strips}
Richard~E Fikes and Nils~J Nilsson.
\newblock Strips: A new approach to the application of theorem proving to
  problem solving.
\newblock \emph{Artificial intelligence}, 2\penalty0 (3-4):\penalty0 189--208,
  1971.

\bibitem[Geva et~al.(2021)Geva, Khashabi, Segal, Khot, Roth, and
  Berant]{geva2021did}
Mor Geva, Daniel Khashabi, Elad Segal, Tushar Khot, Dan Roth, and Jonathan
  Berant.
\newblock Did aristotle use a laptop? a question answering benchmark with
  implicit reasoning strategies.
\newblock \emph{Transactions of the Association for Computational Linguistics},
  9:\penalty0 346--361, 2021.

\bibitem[Goodhart(1984)]{goodhart1984problems}
Charles~AE Goodhart.
\newblock \emph{Problems of monetary management: the UK experience}.
\newblock Springer, 1984.

\bibitem[{Google}(2024)]{google2024gemini}
{Google}.
\newblock Introducing gemini 1.5: Google's next-generation ai model, 2024.
\newblock URL
  \url{https://blog.google/technology/ai/google-gemini-next-generation-model-february-2024/}.
\newblock Accessed: 2024-09-11.

\bibitem[Gu et~al.(2024)Gu, Rozi{\`e}re, Leather, Solar-Lezama, Synnaeve, and
  Wang]{gu2024cruxeval}
Alex Gu, Baptiste Rozi{\`e}re, Hugh Leather, Armando Solar-Lezama, Gabriel
  Synnaeve, and Sida~I Wang.
\newblock Cruxeval: A benchmark for code reasoning, understanding and
  execution.
\newblock \emph{arXiv preprint arXiv:2401.03065}, 2024.

\bibitem[Hendrycks et~al.(2020)Hendrycks, Burns, Basart, Zou, Mazeika, Song,
  and Steinhardt]{hendrycks2020measuring}
Dan Hendrycks, Collin Burns, Steven Basart, Andy Zou, Mantas Mazeika, Dawn
  Song, and Jacob Steinhardt.
\newblock Measuring massive multitask language understanding.
\newblock \emph{arXiv preprint arXiv:2009.03300}, 2020.

\bibitem[Hendrycks et~al.(2021)Hendrycks, Burns, Kadavath, Arora, Basart, Tang,
  Song, and Steinhardt]{hendrycks2021measuring}
Dan Hendrycks, Collin Burns, Saurav Kadavath, Akul Arora, Steven Basart, Eric
  Tang, Dawn Song, and Jacob Steinhardt.
\newblock Measuring mathematical problem solving with the math dataset.
\newblock \emph{arXiv preprint arXiv:2103.03874}, 2021.

\bibitem[Liu et~al.(2024)Liu, Feng, Wang, Wang, Liu, Zhao, Dengr, Ruan, Dai,
  Guo, et~al.]{liu2024deepseek}
Aixin Liu, Bei Feng, Bin Wang, Bingxuan Wang, Bo~Liu, Chenggang Zhao, Chengqi
  Dengr, Chong Ruan, Damai Dai, Daya Guo, et~al.
\newblock Deepseek-v2: A strong, economical, and efficient mixture-of-experts
  language model.
\newblock \emph{arXiv preprint arXiv:2405.04434}, 2024.

\bibitem[McLaughlin(2024)]{mclaughlin2024aidanbench}
Aidan McLaughlin.
\newblock Aidan-bench.
\newblock \url{https://github.com/aidanmclaughlin/Aidan-Bench}, 2024.

\bibitem[Mishra et~al.(2021)Mishra, Khashabi, Baral, and
  Hajishirzi]{mishra2021cross}
Swaroop Mishra, Daniel Khashabi, Chitta Baral, and Hannaneh Hajishirzi.
\newblock Cross-task generalization via natural language crowdsourcing
  instructions.
\newblock \emph{arXiv preprint arXiv:2104.08773}, 2021.

\bibitem[{Mistral AI}(2024{\natexlab{a}})]{mistral2407}
{Mistral AI}.
\newblock Large enough: Mistral large 2 announcement, 2024{\natexlab{a}}.
\newblock URL \url{https://mistral.ai/news/mistral-large-2407/}.
\newblock Accessed: 2024-09-11.

\bibitem[{Mistral AI}(2024{\natexlab{b}})]{mistralnemo2024}
{Mistral AI}.
\newblock Mistral nemo: A state-of-the-art 12b model, 2024{\natexlab{b}}.
\newblock URL \url{https://mistral.ai/news/mistral-nemo/}.
\newblock Accessed: 2024-09-11.

\bibitem[{OpenAI}(2024{\natexlab{a}})]{gpt4o2024}
{OpenAI}.
\newblock Hello gpt-4o, 2024{\natexlab{a}}.
\newblock URL \url{https://openai.com/index/hello-gpt-4o/}.
\newblock Accessed: 2024-09-11.

\bibitem[{OpenAI}(2024{\natexlab{b}})]{gpt4omini2024}
{OpenAI}.
\newblock Gpt-4o mini: Advancing cost-efficient intelligence,
  2024{\natexlab{b}}.
\newblock URL
  \url{https://openai.com/index/gpt-4o-mini-advancing-cost-efficient-intelligence/}.
\newblock Accessed: 2024-09-11.

\bibitem[{OpenAI}(2024{\natexlab{c}})]{openai2024o1system}
{OpenAI}.
\newblock O1 system card, 2024{\natexlab{c}}.
\newblock URL \url{https://cdn.openai.com/o1-system-card.pdf}.
\newblock Accessed: 2024-09-11.

\bibitem[Ouyang et~al.(2022)Ouyang, Wu, Jiang, Almeida, Wainwright, Mishkin,
  Zhang, Agarwal, Slama, Ray, et~al.]{ouyang2022training}
Long Ouyang, Jeffrey Wu, Xu~Jiang, Diogo Almeida, Carroll Wainwright, Pamela
  Mishkin, Chong Zhang, Sandhini Agarwal, Katarina Slama, Alex Ray, et~al.
\newblock Training language models to follow instructions with human feedback.
\newblock \emph{Advances in neural information processing systems},
  35:\penalty0 27730--27744, 2022.

\bibitem[Patel et~al.(2021)Patel, Bhattamishra, and Goyal]{patel2021nlp}
Arkil Patel, Satwik Bhattamishra, and Navin Goyal.
\newblock Are nlp models really able to solve simple math word problems?
\newblock \emph{arXiv preprint arXiv:2103.07191}, 2021.

\bibitem[Recht et~al.(2019)Recht, Roelofs, Schmidt, and
  Shankar]{recht2019imagenet}
Benjamin Recht, Rebecca Roelofs, Ludwig Schmidt, and Vaishaal Shankar.
\newblock Do imagenet classifiers generalize to imagenet?
\newblock In \emph{International conference on machine learning}, pp.\
  5389--5400. PMLR, 2019.

\bibitem[Rein et~al.(2023)Rein, Hou, Stickland, Petty, Pang, Dirani, Michael,
  and Bowman]{rein2023gpqa}
David Rein, Betty~Li Hou, Asa~Cooper Stickland, Jackson Petty, Richard~Yuanzhe
  Pang, Julien Dirani, Julian Michael, and Samuel~R Bowman.
\newblock Gpqa: A graduate-level google-proof q\&a benchmark.
\newblock \emph{arXiv preprint arXiv:2311.12022}, 2023.

\bibitem[Srivastava et~al.(2022)Srivastava, Rastogi, Rao, Shoeb, Abid, Fisch,
  Brown, Santoro, Gupta, Garriga-Alonso, et~al.]{srivastava2022beyond}
Aarohi Srivastava, Abhinav Rastogi, Abhishek Rao, Abu Awal~Md Shoeb, Abubakar
  Abid, Adam Fisch, Adam~R Brown, Adam Santoro, Aditya Gupta, Adri{\`a}
  Garriga-Alonso, et~al.
\newblock Beyond the imitation game: Quantifying and extrapolating the
  capabilities of language models.
\newblock \emph{arXiv preprint arXiv:2206.04615}, 2022.

\bibitem[Talmor et~al.(2018)Talmor, Herzig, Lourie, and
  Berant]{talmor2018commonsenseqa}
Alon Talmor, Jonathan Herzig, Nicholas Lourie, and Jonathan Berant.
\newblock Commonsenseqa: A question answering challenge targeting commonsense
  knowledge.
\newblock \emph{arXiv preprint arXiv:1811.00937}, 2018.

\bibitem[Team et~al.(2024)Team, Riviere, Pathak, Sessa, Hardin, Bhupatiraju,
  Hussenot, Mesnard, Shahriari, Ram{\'e}, et~al.]{team2024gemma}
Gemma Team, Morgane Riviere, Shreya Pathak, Pier~Giuseppe Sessa, Cassidy
  Hardin, Surya Bhupatiraju, L{\'e}onard Hussenot, Thomas Mesnard, Bobak
  Shahriari, Alexandre Ram{\'e}, et~al.
\newblock Gemma 2: Improving open language models at a practical size.
\newblock \emph{arXiv preprint arXiv:2408.00118}, 2024.

\bibitem[Tenenbaum(1999)]{tenenbaum1999rules}
Joshua Tenenbaum.
\newblock Rules and similarity in concept learning.
\newblock \emph{Advances in neural information processing systems}, 12, 1999.

\bibitem[Tenenbaum \& Griffiths(2001)Tenenbaum and
  Griffiths]{tenenbaum2001generalization}
Joshua~B Tenenbaum and Thomas~L Griffiths.
\newblock Generalization, similarity, and bayesian inference.
\newblock \emph{Behavioral and brain sciences}, 24\penalty0 (4):\penalty0
  629--640, 2001.

\bibitem[Tukey(1986)]{tukey1986philosophical}
David~D Tukey.
\newblock A philosophical and empirical analysis of subjects’ modes of
  inquiry in wason's 2--4--6 task.
\newblock \emph{The Quarterly Journal of Experimental Psychology Section A},
  38\penalty0 (1):\penalty0 5--33, 1986.

\bibitem[Tweney et~al.(1980)Tweney, Doherty, Worner, Pliske, Mynatt, Gross, and
  Arkkelin]{tweney1980strategies}
Ryan~D Tweney, Michael~E Doherty, Winifred~J Worner, Daniel~B Pliske,
  Clifford~R Mynatt, Kimberly~A Gross, and Daniel~L Arkkelin.
\newblock Strategies of rule discovery in an inference task.
\newblock \emph{Quarterly Journal of Experimental Psychology}, 32\penalty0
  (1):\penalty0 109--123, 1980.

\bibitem[Valmeekam et~al.(2022)Valmeekam, Olmo, Sreedharan, and
  Kambhampati]{valmeekam2022large}
Karthik Valmeekam, Alberto Olmo, Sarath Sreedharan, and Subbarao Kambhampati.
\newblock Large language models still can't plan (a benchmark for llms on
  planning and reasoning about change).
\newblock In \emph{NeurIPS 2022 Foundation Models for Decision Making
  Workshop}, 2022.

\bibitem[Valmeekam et~al.(2024)Valmeekam, Stechly, and
  Kambhampati]{valmeekam2024llmscantplanlrms}
Karthik Valmeekam, Kaya Stechly, and Subbarao Kambhampati.
\newblock Llms still can't plan; can lrms? a preliminary evaluation of openai's
  o1 on planbench, 2024.
\newblock URL \url{https://arxiv.org/abs/2409.13373}.

\bibitem[Vaswani(2017)]{vaswani2017attention}
Ashish Vaswani.
\newblock Attention is all you need.
\newblock \emph{Advances in Neural Information Processing Systems}, 2017.

\bibitem[Wang et~al.(2024)Wang, Wang, Athiwaratkun, Zhang, and
  Zou]{wang2024mixture}
Junlin Wang, Jue Wang, Ben Athiwaratkun, Ce~Zhang, and James Zou.
\newblock Mixture-of-agents enhances large language model capabilities.
\newblock \emph{arXiv preprint arXiv:2406.04692}, 2024.

\bibitem[Wang et~al.(2023{\natexlab{a}})Wang, Zelikman, Poesia, Pu, Haber, and
  Goodman]{wang2023hypothesis}
Ruocheng Wang, Eric Zelikman, Gabriel Poesia, Yewen Pu, Nick Haber, and Noah~D
  Goodman.
\newblock Hypothesis search: Inductive reasoning with language models.
\newblock \emph{arXiv preprint arXiv:2309.05660}, 2023{\natexlab{a}}.

\bibitem[Wang et~al.(2023{\natexlab{b}})Wang, Hu, Lu, Zhu, Zhang, Subramaniam,
  Loomba, Zhang, Sun, and Wang]{wang2023scibench}
Xiaoxuan Wang, Ziniu Hu, Pan Lu, Yanqiao Zhu, Jieyu Zhang, Satyen Subramaniam,
  Arjun~R Loomba, Shichang Zhang, Yizhou Sun, and Wei Wang.
\newblock Scibench: Evaluating college-level scientific problem-solving
  abilities of large language models.
\newblock \emph{arXiv preprint arXiv:2307.10635}, 2023{\natexlab{b}}.

\bibitem[Wang et~al.(2023{\natexlab{c}})Wang, Wang, Liu, Chen, Yuan, Peng, and
  Ji]{wang2023mint}
Xingyao Wang, Zihan Wang, Jiateng Liu, Yangyi Chen, Lifan Yuan, Hao Peng, and
  Heng Ji.
\newblock Mint: Evaluating llms in multi-turn interaction with tools and
  language feedback.
\newblock \emph{arXiv preprint arXiv:2309.10691}, 2023{\natexlab{c}}.

\bibitem[Wason(1960)]{wason1960failure}
Peter~C Wason.
\newblock On the failure to eliminate hypotheses in a conceptual task.
\newblock \emph{Quarterly journal of experimental psychology}, 12\penalty0
  (3):\penalty0 129--140, 1960.

\bibitem[Wei et~al.(2022)Wei, Wang, Schuurmans, Bosma, Xia, Chi, Le, Zhou,
  et~al.]{wei2022chain}
Jason Wei, Xuezhi Wang, Dale Schuurmans, Maarten Bosma, Fei Xia, Ed~Chi, Quoc~V
  Le, Denny Zhou, et~al.
\newblock Chain-of-thought prompting elicits reasoning in large language
  models.
\newblock \emph{Advances in neural information processing systems},
  35:\penalty0 24824--24837, 2022.

\bibitem[Yang et~al.(2024{\natexlab{a}})Yang, Jimenez, Wettig, Lieret, Yao,
  Narasimhan, and Press]{yang2024swe}
John Yang, Carlos~E Jimenez, Alexander Wettig, Kilian Lieret, Shunyu Yao,
  Karthik Narasimhan, and Ofir Press.
\newblock Swe-agent: Agent-computer interfaces enable automated software
  engineering.
\newblock \emph{arXiv preprint arXiv:2405.15793}, 2024{\natexlab{a}}.

\bibitem[Yang et~al.(2024{\natexlab{b}})Yang, Yang, Gao, Wang, Wang, Zhu, Chu,
  Zhou, Liang, Wang, et~al.]{yang2024empirical}
Lin Yang, Chen Yang, Shutao Gao, Weijing Wang, Bo~Wang, Qihao Zhu, Xiao Chu,
  Jianyi Zhou, Guangtai Liang, Qianxiang Wang, et~al.
\newblock An empirical study of unit test generation with large language
  models.
\newblock \emph{arXiv preprint arXiv:2406.18181}, 2024{\natexlab{b}}.

\bibitem[Yang et~al.(2023)Yang, Chiang, Zheng, Gonzalez, and
  Stoica]{yang2023rethinking}
Shuo Yang, Wei-Lin Chiang, Lianmin Zheng, Joseph~E Gonzalez, and Ion Stoica.
\newblock Rethinking benchmark and contamination for language models with
  rephrased samples.
\newblock \emph{arXiv preprint arXiv:2311.04850}, 2023.

\bibitem[Zhu et~al.(2024)Zhu, Guo, Shao, Yang, Wang, Xu, Wu, Li, Gao, Ma,
  et~al.]{zhu2024deepseek}
Qihao Zhu, Daya Guo, Zhihong Shao, Dejian Yang, Peiyi Wang, Runxin Xu, Y~Wu,
  Yukun Li, Huazuo Gao, Shirong Ma, et~al.
\newblock Deepseek-coder-v2: Breaking the barrier of closed-source models in
  code intelligence.
\newblock \emph{arXiv preprint arXiv:2406.11931}, 2024.

\end{thebibliography}
\bibliographystyle{iclr2025_conference}

\appendix

\section{Appendix / Supplemental Material}

\subsection{Experimental Details}
\label{appendix:experimental-details}

\subsubsection{Model Configurations}

Models are accessed when possible with default configurations and settings provided by their respective proprietary APIs (i.e. Anthropic, OpenAI, Mistral, DeepSeek, and Vertex APIs). Llama 3.1 models are run via the Amazon Bedrock API. Gemma 2 9b and llama 3 70B are run via the Groq API \cite{abts2022software}. 

\subsubsection{Prompts}

All models begin from the same initial prompt. 

\begin{quotation}
    You are an AI tasked with discovering a hidden rule that takes three numbers as input and returns a boolean. These numbers can be any float or integer number that can be typed in 3 or fewer digits, not including signs and decimal characters (e.g. -11.3 is a valid value). In each response, you will provide your step-by-step reasoning for your future plans on this task, and then you will take one of two actions:

    \begin{enumerate}
        \item Provide a single test case to observe the output of a rule (e.g. ``\verb|Test Case: ```(1, 2, 3)```|'' which corresponds to x=1, y=2, z=3. You may only propose one test case at a time.
        \item Provide a final ``guess'' in the form of a Python lambda function (e.g. ``\verb|Final Guess: ```lambda x,y,z: x < y < z```|''). You only get to select this action one time, so don't use it unless you are confident or are forced to.
    \end{enumerate}

    You will have up to 30 attempts at test cases, and you will have exactly one attempt at guessing the final function. Make sure all of the tests you provide are unique from other ones you have already tried -- the functions are all deterministic, and will be the same every time. You should prioritize getting the rule as correct as possible, since an incorrect answer will get no points. You should therefore not guess the function unless you are quite confident, or if you are very close to running out of attempts.

    You must return your responses in the format laid out above at the very bottom of your message. For example, if you want to submit a test case, you must conclude with the string ``\verb|Test Case: ```(x,y,z)```|'', where x,y,z are replaced with your guesses. If you want to submit a final guess, you must conclude with the string ``\verb|Final Guess: ```<function>```|'' where \verb|<function>| is replaced with a Python lambda function. Do not include any comments or additional text on the same lines as these two things.

    Make sure to include your reasoning for your tests -- what you are testing for, why you selected that test, etc.
\end{quotation}

Responses by the models are pulled out via regular expressions matching the formatting in the prompt. We find that after controlling for various formatting eccentricities (Python blocks, markdown, bold characters, etc) that all listed models are capable of providing test cases in this format.

\subsubsection{Verifying Equivalent Functions}

To verify two provided lambda functions are equivalent, we generate a large number of test cases and ensure the provided rules match on all outputs. Specifically, we create three sets of cases:

\begin{enumerate}
    \item \textbf{Integer Grid Cases} - We construct a 40x40x40 grid of integer triplets from -20 to 20, inclusive, leading to 64,000 triplet cases.
    \item \textbf{Random Uniform Cases} - We construct a list of 10,000 uniformly random float triplets from -200 to 200, inclusive.
    \item \textbf{Special Cases} - We hand-design a small set of test cases to ensure all hidden rules in the full split are adequately tested.
\end{enumerate}

We mark a rule as incorrect if any test cases generated above show different behavior between the hidden rule and the guessed rule, and mark it correct otherwise. 

\subsection{Test Case Novelty}
\label{appendix:novelty}

Test case novelty is an interesting second order metric for success upon the WILT task. Broadly speaking, models that reuse fewer tests are rewarded with more information for which to solve the task. Models that very rarely re-propose a test tend to perform very well upon WILT, but the inverse is not necessarily true -- models that loop tests often still arrive at the right answer.

Repeated tests are useful for bifurcating the types of failures on WILT -- one being the doom loop phenomenon, and the other being reasoning capability conditioned upon some available evidence. One hypothesis for the observed behavior is that certain models are primarily oriented towards single-turn scenarios, and that one type of failure need not imply the other. DeepSeek Chat v2.5, for example, demonstrates strong initial hypothesis space reduction compared to other models, which allows it notably better performance on WILT compared to other models with  similar repeat counts (e.g. open-mistral-nemo). Strong single-turn performance and deductive reasoning capabilities can help salvage performance from a model that demonstrates difficulty with multi-turn inductive logic.


\begin{table}
  \caption{Model Novelty Metrics}
  \label{tab:model-novelty}
  \centering
  \begin{tabular}{llll}
    \toprule
    \textbf{Model} & \textbf{Repeats} & \textbf{Avg. Novelty} & \textbf{Avg. Min. Novelty} \\
    \cmidrule(r){1-4}
    claude-3.5-sonnet-20240620 & 27 & 0.22 & 0.08 \\
    o1-mini-2024-09-12 & \textbf{3} & 0.24 & 0.09 \\
    o1-preview-2024-09-12 & \textbf{3} & \textbf{0.28} & \textbf{0.11} \\
    chatgpt-4o-latest & 38 & 0.21 & 0.10 \\
    mistral-large-2407 & 142 & 0.13 & 0.04 \\
    gpt-4o-2024-08-06 & 26 & 0.22 & 0.11 \\
    llama3.1 405B & 30 & 0.20 & 0.06 \\
    gemini-1.5-flash-exp-0827 & 108 & 0.19 & 0.08 \\
    llama3-70b & 74 & 0.19 & 0.06 \\
    deepseek-chat-v2.5 & 489 & 0.09 & 0.02 \\
    gpt-4o-mini & 54 & 0.19 & 0.10 \\
    gemini-1.5-pro & 41 & 0.26 & 0.13 \\
    gemini-1.5-flash & 123 & 0.19 & 0.08 \\
    deepseek-coder & 334 & 0.11 & 0.03 \\
    deepseek-chat-v2 & 335 & 0.10 & 0.03 \\
    llama-3.1-8b & 223 & 0.13 & 0.01 \\
    open-mistral-nemo & 400 & 0.12 & 0.02 \\
    claude-3-haiku-20240307 & 11 & 0.29 & 0.11 \\
    gemini-1.5-flash-8b-exp-0827 & 386 & 0.14 & 0.05 \\
    gemma2-9b-it & 70 & 0.34 & 0.18 \\
    \bottomrule
  \end{tabular}
\end{table}

Following Aidan-bench \citep{mclaughlin2024aidanbench}, we provide Table \ref{tab:model-novelty}, containing additional novelty metrics. These include: 

\begin{enumerate}
  \item \textit{Average novelty} - which reports the average cosine similarity between each message's gpt-3 embeddings and the closest previous message within the same test. 
  \item \textit{Average minimum novelty} - which reports the average \textit{minimum} cosine similarity between each message's gpt-3 embeddings and the closest previous message within the same test.
\end{enumerate}

\begin{figure} [ht]
    \centering
    \includegraphics[width=1\linewidth]{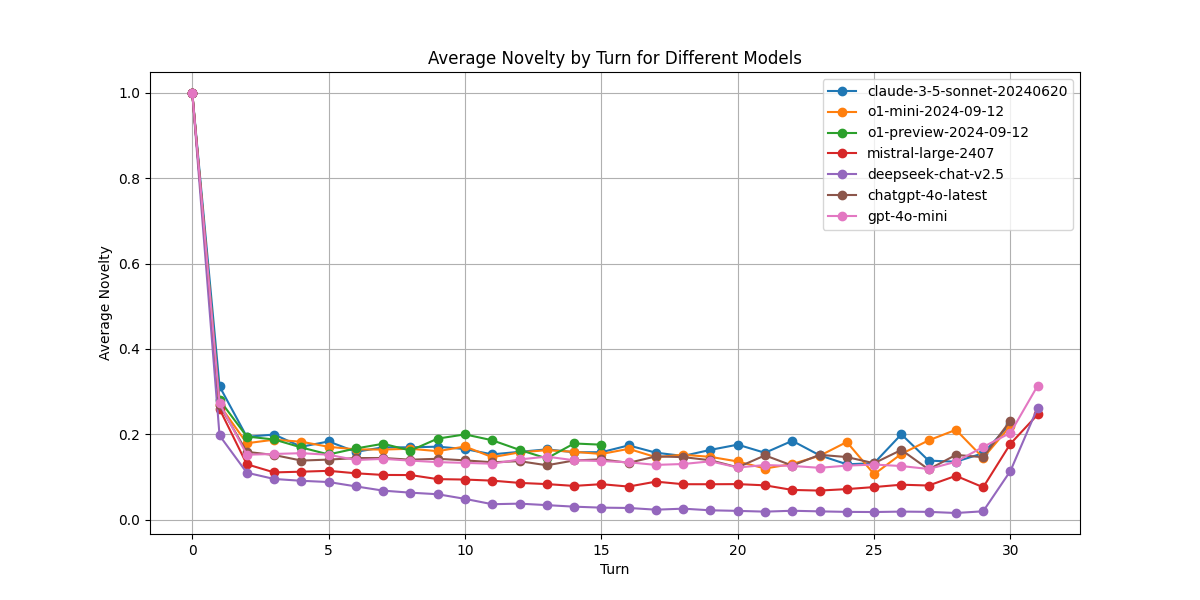}
    \caption{Cosine similarity by turn for selected models. Models have a higher novelty score near the end, since the final guess is often much different from previous messages, which are all proposed test cases.}
    \label{fig:novelty-turn}
\end{figure}

These capture an additional dimension of ``test novelty'' and ``message novelty'', where models may propose the same tests for different reasons, or repeat previously generated messages verbatim. We bold results which are best within the class of high performing models. We note that models that propose fewer tests before guessing (e.g. o1-preview, o1-mini) should see lower values for all of these compared to models that tend to use many tests before guessing (e.g. mistral-large) even for otherwise equally performing models. We also show the average novelty by turn in Figure \ref{fig:novelty-turn}, which captures the model's novelty scores across turns. 

\subsection{Response Complexity in Test Swap Experiments}
\label{appendix:testswapcomplex}

\begin{figure} [ht]
    \centering
    \includegraphics[width=0.8\linewidth]{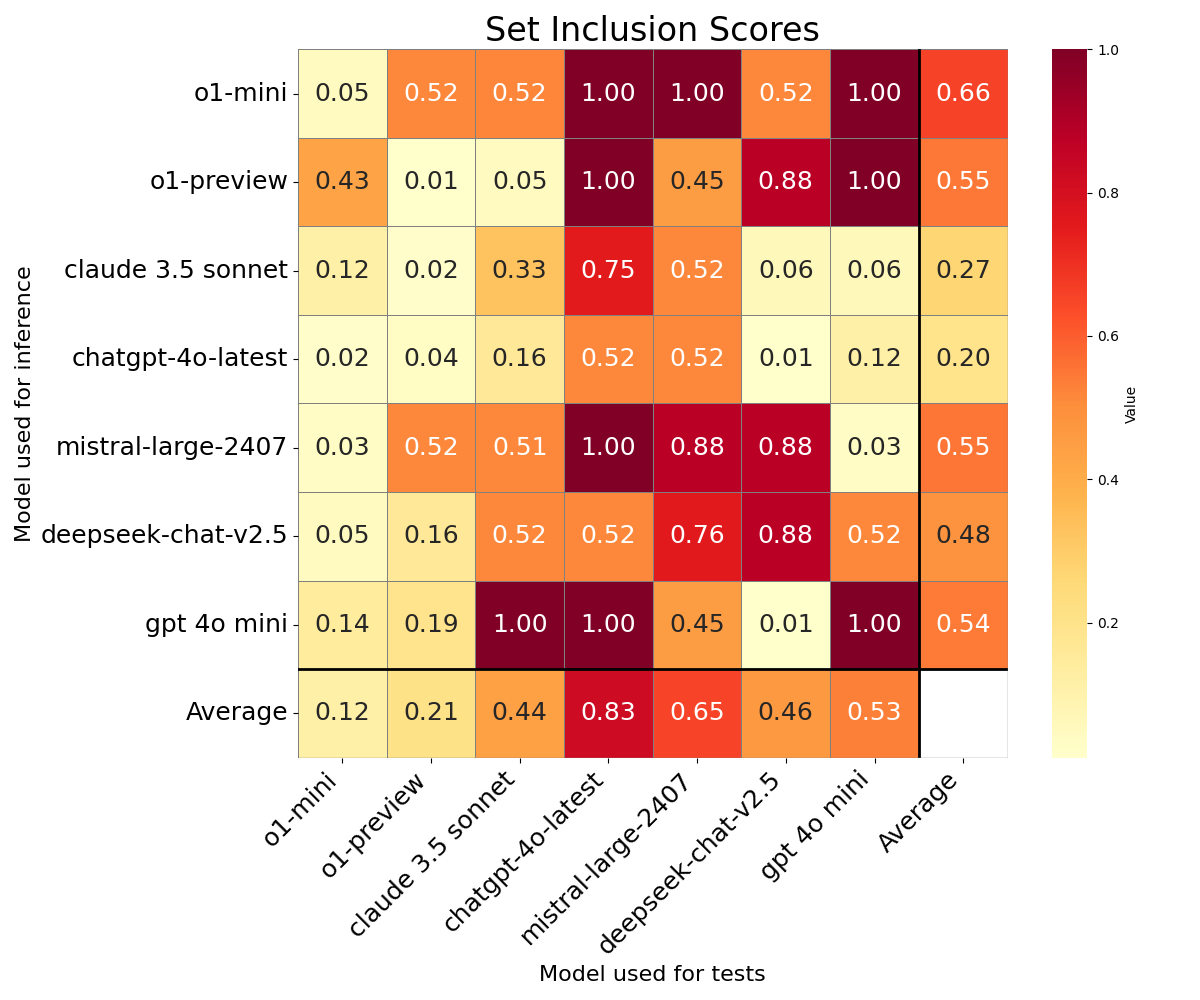}
    \caption{Set inclusion ratios will differ in the same model when provided another model's tests, and models provided the same tests have different set inclusion behaviors.}
    \label{fig:setinclusion}
\end{figure}

In Figure \ref{fig:setinclusion} we show the set inclusion ratios in the case where a model is provided another model's test cases. That is, we show whether an error in the final guess of a model is likely to be smaller / less than one (e.g. $x < y < z$ instead of $x \leq y \leq z$), or larger / greater than one (e.g. $x > 0$ instead of $x > 0 \land x < 5$). This seems more test-case dependent than other complexity benchmarks, where tests provided by certain models seem to lead to smaller hypotheses.

\begin{figure} [ht]
    \centering
    \includegraphics[width=0.8\linewidth]{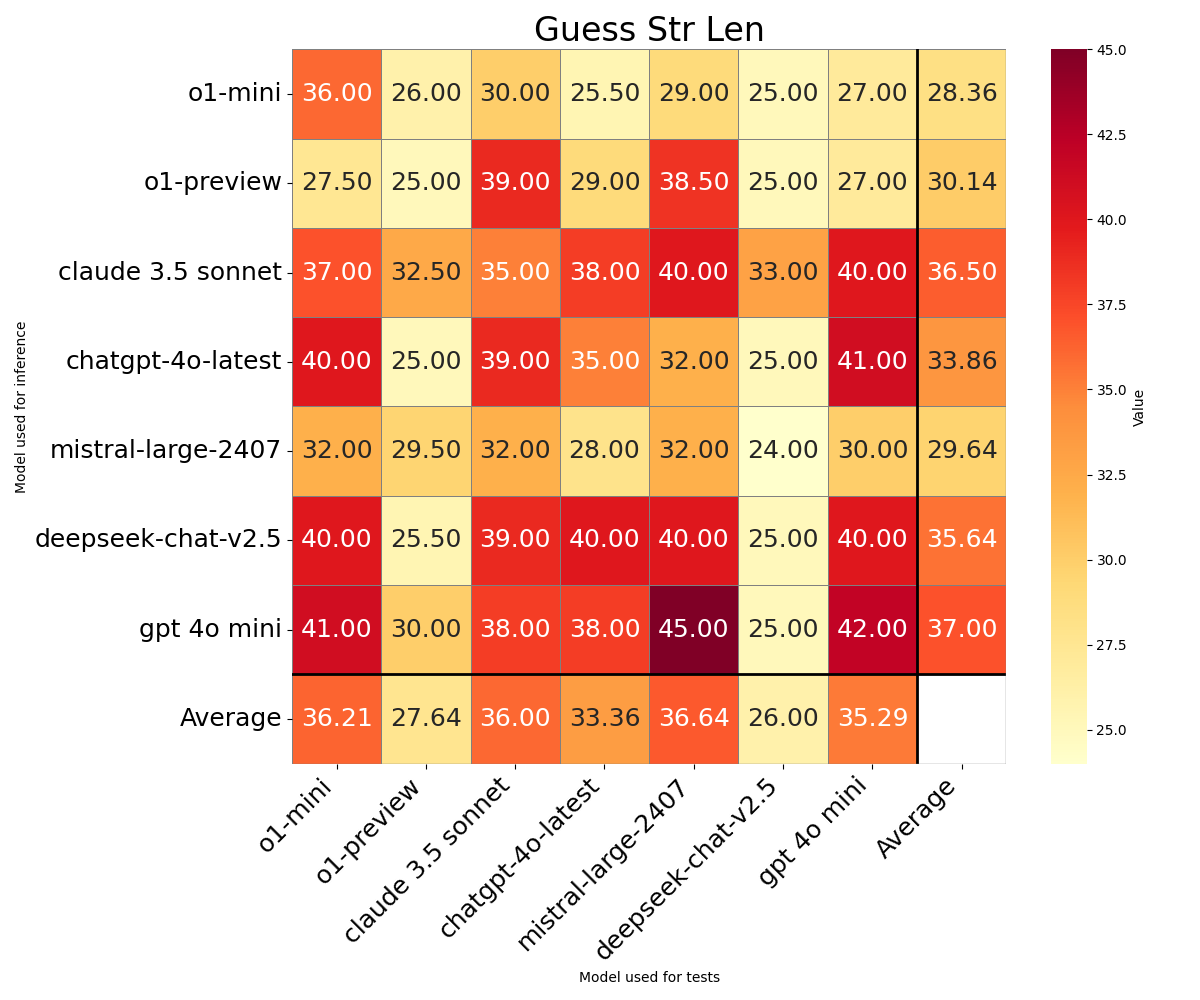}
    \caption{Models tend to have fairly consistent guess string lengths, with some exceptions.}
    \label{fig:strlen_swap}
\end{figure}

Figure \ref{fig:strlen_swap} shows the length of the string used to guess the rule by each model, which is comparatively more consistent for a model. We find this to be fairly consistent across settings, with most models hovering near 45.

\newpage

\subsection{Selected Lite Split Results}
\label{appendix:lite-split}

Table \ref{tab:lite-split} contains selected results for the \textit{lite split}. We show this much easier split produces a similar ordering, suggesting that the bulk of the separation in our benchmark lies in the easier tests.

\subsection{Test Descriptions}

Table \ref{tab:tests_lite} contains the functions found in the \textit{lite split} of WILT. Table \ref{tab:tests_full} contains the functions found in the \textit{full split} of WILT. These are implemented as Python lambda functions. 

\begin{table}
  \caption{Lite Split Metrics}
  \label{tab:lite-split}
  \centering
  \begin{tabular}{lcc}
    \toprule
    \textbf{Model} & \textbf{Accuracy} & \textbf{Avg. Guesses} \\
    \midrule
        Claude-3.5-Sonnet & 8/10 & 13.68 \\
        GPT-4-Turbo & 7/10 & 12.47 \\
        GPT-4o & 6/10 & 13.86 \\
        DeepSeek-V2-Coder & 6/10 & 23.46 \\
        Llama 3 70B & 4/10 & 15.64 \\
        DeepSeek-V2-Chat & 2/10 & 24.82 \\
        Llama 3 8B & 1/10 & 24.0 \\
        GPT-3.5-Turbo & 1/10 & 2.9 \\
    \bottomrule
  \end{tabular}
\end{table}

\begin{table}
  \caption{Lite Split: Complete Set of Tests}
  \label{tab:tests_lite}
  \centering
  \begin{tabular}{cl}
  \toprule
    \textbf{Test ID} & \textbf{Description} \\
    \cmidrule(r){1-2}
        1 & $x > y > z$ \\
        2 & $x < y < z$ \\
        3 & $x \geq y \geq z$ \\
        4 & $x \leq y \leq z$ \\
        5 & $x = y = z$ \\
        6 & $x \neq y \land y \neq z \land x \neq z$ \\
        7 & $x < 0 \land y < 0 \land z < 0$ \\
        8 & $x + y = z$ \\
        9 & $x \cdot y = z$ \\
        10 & $x < y \land y > z$ \\
    \bottomrule
  \end{tabular}
\end{table}

\begin{table}
  \caption{Full Split: Complete Set of Tests}
  \label{tab:tests_full}
  \centering
  \begin{tabular}{cl}
    \toprule
    \textbf{Test ID} & \textbf{Description} \\
    \cmidrule(r){1-2}
    \multicolumn{2}{c}{\textbf{Easy Tests}} \\
    \cmidrule(r){1-2}
    1 & $x > y > z$ \\
    2 & $x < y < z$ \\
    3 & $x \geq y \geq z$ \\
    4 & $x \leq y \leq z$ \\
    5 & $x < z < y$ \\
    6 & $x \leq z \leq y$ \\
    7 & $z < x < y$ \\
    8 & $z \leq x \leq y$ \\
    9 & $x = y = z$ \\
    10 & $x \neq y \land y \neq z \land x \neq z$ \\
    11 & $x < 0 \land y < 0 \land z < 0$ \\
    12 & $x > 0 \land y > 0 \land z > 0$ \\
    13 & $x \bmod 2 = 0 \land y \bmod 2 = 0 \land z \bmod 2 = 0$ \\
    14 & $x \bmod 2 \neq 0 \land y \bmod 2 \neq 0 \land z \bmod 2 \neq 0$ \\
    \cmidrule(r){1-2}
    \multicolumn{2}{c}{\textbf{Medium Tests}} \\
    \cmidrule(r){1-2}
    15 & $x + y = z$ \\
    16 & $x \cdot y = z$ \\
    17 & $x + z = y$ \\
    18 & $x \cdot z = y$ \\
    19 & $y + z = x$ \\
    20 & $y \cdot z = x$ \\
    21 & $\max(x, y, z) = x$ \\
    22 & $\max(x, y, z) = y$ \\
    23 & $\max(x, y, z) = z$ \\
    24 & $\min(x, y, z) = x$ \\
    25 & $\min(x, y, z) = y$ \\
    26 & $\min(x, y, z) = z$ \\
    27 & $x + y + z = 0$ \\
    28 & $x \cdot y \cdot z = 0$ \\
    29 & $(x + y + z) \bmod 2 = 0$ \\
    30 & $(x + y + z) \bmod 2 = 1$ \\
    31 & $(x \cdot y \cdot z) \bmod 2 = 0$ \\
    32 & $(x \cdot y \cdot z) \bmod 2 = 1$ \\
    33 & $(x + y) / 2 = z$ \\
    34 & $-5 \leq x, y, z \leq 5$ \\
    35 & $-10 \leq x, y, z \leq 10$ \\
    36 & $-5 \leq x, y, z \leq 0$ \\
    37 & $0 \leq x, y, z \leq 5$ \\
    38 & $-2 \leq x, y, z \leq 2$ \\
    39 & $-20 \leq x, y, z \leq 20$ \\
    \cmidrule(r){1-2}
    \multicolumn{2}{c}{\textbf{Very Hard Tests}} \\
    \cmidrule(r){1-2}
    40 & $x^2 + y^2 = z^2$ \\
    41 & $x^2 + z^2 = y^2$ \\
    42 & $y^2 + z^2 = x^2$ \\
    43 & $\lfloor x \rfloor \& \lfloor y \rfloor = \lfloor z \rfloor$ (bitwise AND) \\
    44 & $\lfloor x \rfloor | \lfloor y \rfloor = \lfloor z \rfloor$ (bitwise OR) \\
    45 & $\lfloor x \rfloor \oplus \lfloor y \rfloor = \lfloor z \rfloor$ (bitwise XOR) \\
    46 & $x, y, z \in \mathbb{Z} \land \gcd(x, y) = \gcd(y, z) = \gcd(z, x) = 1$ \\
    47 & $\sqrt{\lfloor |x| \rfloor}^2 = \lfloor |x| \rfloor \land \sqrt{\lfloor |y| \rfloor}^2 = \lfloor |y| \rfloor \land \sqrt{\lfloor |z| \rfloor}^2 = \lfloor |z| \rfloor$ \\
    48 & $0 < x \bmod 1 \land 0 < y \bmod 1 \land 0 < z \bmod 1$ \\
    49 & $0 < x \bmod 1 < y \bmod 1 < z \bmod 1 < 1$ \\
    50 & $x < y < z \land 0 < z - x \leq 1$ \\
    \bottomrule
  \end{tabular}
\end{table}

\newpage

\section{Failure Case Examples}
\label{appendix:failures}

\subsection{Doom Loop Without Reasoning Failure}

Table \ref{tab:deepseek-loop} shows an example of a model (Deepseek-Chat-v2.5) entering a doom loop during the test proposal phase, but where that doom loop does not constitute a reasoning failure. Reasoning has been removed for brevity.

\begin{table}
  \caption{DeepSeek-v2.5 Enters a doom loop but still arrives at the correct answer}
  \label{tab:deepseek-loop}
\begin{tabular}{cp{6cm}p{5cm}}
\cmidrule(r){1-3}
\textbf{Turn} & \textbf{Assistant} & \textbf{Harness} \\
\cmidrule(r){1-3}
1 & Test Case: (1, 1, 1) & (1.0, 1.0, 1.0): True. \\
\cmidrule(r){1-3}
2 & Test Case: (1, 2, 3) & (1.0, 2.0, 3.0): True. \\
\cmidrule(r){1-3}
3 & Test Case: (3, 2, 1) & (3.0, 2.0, 1.0): True. \\
\cmidrule(r){1-3}
4 & Test Case: (2, 1, 3) & (2.0, 1.0, 3.0): True. \\
\cmidrule(r){1-3}
5 & Test Case: (-1, -2, -3) & (-1.0, -2.0, -3.0): False. \\
\cmidrule(r){1-3}
6 & Test Case: (-1, 2, -3) & (-1.0, 2.0, -3.0): False. \\
\cmidrule(r){1-3}
7 & Test Case: (0, 1, 2) & (0.0, 1.0, 2.0): False. \\
\cmidrule(r){1-3}
8 & Test Case: (2, 3, 1) & (2.0, 3.0, 1.0): True. \\
\cmidrule(r){1-3}
9 & Test Case: (0, 1, 1) & (0.0, 1.0, 1.0): False. \\
\cmidrule(r){1-3}
10 & Test Case: (0.001, 1, 1) & (0.001, 1.0, 1.0): True. \\
\cmidrule(r){1-3}
11 & Test Case: (-0.001, 1, 1) & (-0.001, 1.0, 1.0): False. \\
\cmidrule(r){1-3}
12 & Test Case: (999, 1, 1) & (999.0, 1.0, 1.0): True. \\
\cmidrule(r){1-3}
13 & Test Case: (-999, 1, 1) & (-999.0, 1.0, 1.0): False. \\
\cmidrule(r){1-3}
14 & Test Case: (0.0001, 1, 1) & (0.0001, 1.0, 1.0): True. \\
\cmidrule(r){1-3}
15 & Test Case: (-0.0001, 1, 1) & (-0.0001, 1.0, 1.0): False. \\
\cmidrule(r){1-3}
16 & Test Case: (999.999, 1, 1) & (999.999, 1.0, 1.0): True. \\
\cmidrule(r){1-3}
17 & Test Case: (-999.999, 1, 1) & (-999.999, 1.0, 1.0): False. \\
\cmidrule(r){1-3}
18 & Test Case: (0.0001, 999.999, 1) & (0.0001, 999.999, 1.0): True. \\
\cmidrule(r){1-3}
19 & Test Case: (-0.0001, 999.999, 1) & (-0.0001, 999.999, 1.0): False. \\
\cmidrule(r){1-3}
20 & Test Case: (0.0001, -999.999, 1) & (0.0001, -999.999, 1.0): False. \\
\cmidrule(r){1-3}
21 & Test Case: (0.0001, 999.999, -0.0001) & (0.0001, 999.999, -0.0001): False. \\
\cmidrule(r){1-3}
22 & Test Case: (0.0001, 999.999, -999.999) & (0.0001, 999.999, -999.999): False. \\
\cmidrule(r){1-3}
23 & Test Case: (0.0001, 999.999, 0.0001) & (0.0001, 999.999, 0.0001): True. \\
\cmidrule(r){1-3}
24 & Test Case: (0.0001, 0.0001, 999.999) & (0.0001, 0.0001, 999.999): True. \\
\cmidrule(r){1-3}
25 & Test Case: (999.999, 0.0001, 0.0001) & (999.999, 0.0001, 0.0001): True. \\
\cmidrule(r){1-3}
26 & Test Case: (0.0001, 999.999, 0.0001) & (0.0001, 999.999, 0.0001): True. \\
\cmidrule(r){1-3}
27 & Test Case: (0.0001, 0.0001, 999.999) & (0.0001, 0.0001, 999.999): True. \\
\cmidrule(r){1-3}
28 & Test Case: (999.999, 0.0001, 0.0001) & (999.999, 0.0001, 0.0001): True. \\
\cmidrule(r){1-3}
29 & Test Case: (0.0001, 999.999, 0.0001) & (0.0001, 999.999, 0.0001): True. \\
\cmidrule(r){1-3}
30 & Test Case: (0.0001, 0.0001, 999.999) & (0.0001, 0.0001, 999.999): True. \\
\cmidrule(r){1-3}
31 & Final Guess: lambda x, y, z: $x > 0$ and $y > 0$ and $z > 0$ & Congratulations! Your guess is correct. \\
\cmidrule(r){1-3}
\end{tabular}
\end{table}

\subsection{Approximately Correct}

Table \ref{tab:o1-mini-test-cases} shows an example of o1-mini getting a very difficult test case (co-primality) approximately correct, failing only because it adds an additional arbitrary constraint upon the magnitude of the values despite no such constraint existing. Reasoning has been removed for brevity.

\begin{table}
\caption{o1-mini nearly gets the coprimality case correct, but adds bounds for no concrete reason.}
\label{tab:o1-mini-test-cases}
\begin{tabular}{cp{6cm}p{5cm}}
\cmidrule(r){1-3}
\textbf{Turn} & \textbf{Assistant} & \textbf{Harness} \\
\cmidrule(r){1-3}
1 & Test Case: (1, 2, 3) & (1.0, 2.0, 3.0): True. \\
\cmidrule(r){1-3}
2 & Test Case: (3, 2, 1) & (3.0, 2.0, 1.0): True. \\
\cmidrule(r){1-3}
3 & Test Case: (1, 1, 1) & (1.0, 1.0, 1.0): True. \\
\cmidrule(r){1-3}
4 & Test Case: (-1, 2, 3) & (-1.0, 2.0, 3.0): True. \\
\cmidrule(r){1-3}
5 & Test Case: (-1, -2, -3) & (-1.0, -2.0, -3.0): True. \\
\cmidrule(r){1-3}
6 & Test Case: (0, 0, 0) & (0.0, 0.0, 0.0): False. \\
\cmidrule(r){1-3}
7 & Test Case: (0, 1, 2) & (0.0, 1.0, 2.0): False. \\
\cmidrule(r){1-3}
8 & Test Case: (1.5, 2.5, 3.5) & (1.5, 2.5, 3.5): False. \\
\cmidrule(r){1-3}
9 & Test Case: (1, 2, 3.0) & (1.0, 2.0, 3.0): True. \\
\cmidrule(r){1-3}
10 & Test Case: (1, 2, 3.1) & (1.0, 2.0, 3.1): False. \\
\cmidrule(r){1-3}
11 & Test Case: (1, 2, -3) & (1.0, 2.0, -3.0): True. \\
\cmidrule(r){1-3}
12 & Test Case: (10, -20, 30) & (10.0, -20.0, 30.0): False. \\
\cmidrule(r){1-3}
13 & Test Case: (4, 5, 6) & (4.0, 5.0, 6.0): False. \\
\cmidrule(r){1-3}
14 & Test Case: (-2, -1, 3) & (-2.0, -1.0, 3.0): True. \\
\cmidrule(r){1-3}
15 & Test Case: (3, 3, 4) & (3.0, 3.0, 4.0): False. \\
\cmidrule(r){1-3}
16 & Test Case: (1, -1, 2) & (1.0, -1.0, 2.0): True. \\
\cmidrule(r){1-3}
17 & Test Case: (-3, 3, -2) & (-3.0, 3.0, -2.0): False. \\
\cmidrule(r){1-3}
18 & Test Case: (2, -2, 2) & (2.0, -2.0, 2.0): False. \\
\cmidrule(r){1-3}
19 & Test Case: (-1, 2, -2) & (-1.0, 2.0, -2.0): False. \\
\cmidrule(r){1-3}
20 & Test Case: (1, 3, 2) & (1.0, 3.0, 2.0): True. \\
\cmidrule(r){1-3}
21 & Test Case: (1, 2, 2) & (1.0, 2.0, 2.0): False. \\
\cmidrule(r){1-3}
22 & Test Case: (-1, -2, 3) & (-1.0, -2.0, 3.0): True. \\
\cmidrule(r){1-3}
23 & Test Case: (2, 3, -1) & (2.0, 3.0, -1.0): True. \\
\cmidrule(r){1-3}
24 & Final Guess: lambda x, y, z: all(isinstance(n, int) and $n \neq 0$ and $-3 \leq n \leq 3$ for n in [x, y, z]) and math.gcd(x, y) == 1 and math.gcd(x, z) == 1 and math.gcd(y, z) == 1 & Sorry, that's not the correct rule. \\
\cmidrule(r){1-3}
\end{tabular}
\end{table}

\subsection{Confirmation Bias}
\label{conf-bias-example}

Table \ref{tab:o1-preview-confirmation} shows an example of o1-preview failing a relatively easy test case $(x \geq y \geq z)$ due to a confirmation bias error. The model uses only 9 test cases and correctly identifies that the rule returns true when all three are equal, but submits five test cases confirming that and none exploring other rules which are true when three items are equal. Reasoning has been removed for brevity.

\begin{table}
\caption{o1-preview fails early on an easy test case despite having a wealth of remaining tests, because it latched on to an early behavior and sought to confirm it rather than falsify it.}
\label{tab:o1-preview-confirmation}
\begin{tabular}{cp{6cm}p{5cm}}
\cmidrule(r){1-3}
\textbf{Turn} & \textbf{Assistant} & \textbf{Harness} \\
\cmidrule(r){1-3}
1 & Test Case: (2, 3, 6) & (2.0, 3.0, 6.0): False. \\
\cmidrule(r){1-3}
2 & Test Case: (2, 3, 5) & (2.0, 3.0, 5.0): False. \\
\cmidrule(r){1-3}
3 & Test Case: (1, 1, 1) & (1.0, 1.0, 1.0): True. \\
\cmidrule(r){1-3}
4 & Test Case: (2, 2, 2) & (2.0, 2.0, 2.0): True. \\
\cmidrule(r){1-3}
5 & Test Case: (1, 1, 2) & (1.0, 1.0, 2.0): False. \\
\cmidrule(r){1-3}
6 & Test Case: (-1, -1, -1) & (-1.0, -1.0, -1.0): True. \\
\cmidrule(r){1-3}
7 & Test Case: (1.5, 1.5, 1.5) & (1.5, 1.5, 1.5): True. \\
\cmidrule(r){1-3}
8 & Test Case: (0, 0, 0) & (0.0, 0.0, 0.0): True. \\
\cmidrule(r){1-3}
9 & Test Case: (1, 2, 3) & (1.0, 2.0, 3.0): False. \\
\cmidrule(r){1-3}
10 & Final Guess: lambda x, y, z: x == y == z & Sorry, that's not the correct rule. \\
\cmidrule(r){1-3}
\end{tabular}
\end{table}

\subsection{Same Test For New Reason}

Table \ref{tab:new-reason} shows an example of Claude Sonnet 3.5 repeating a test, where it will mistakenly generate the same test for a different stated reason. We see the model notice it has repeated a test only after it has already submitted the test. Other tests have been removed for brevity.

\begin{table}
\caption{Claude Sonnet 3.5 double submits $(2,3,5)$ to test $x + y = z$ and $all(isprime(n))$}
\label{tab:new-reason}
\begin{tabular}{cp{6cm}p{5cm}}
\cmidrule(r){1-3}
\textbf{Turn} & \textbf{Assistant} & \textbf{Harness} \\
\cmidrule(r){1-3}
1-5 & ... & ... \\
\cmidrule(r){1-3}
6 & Still False. Let's try a different approach and consider some mathematical relationships. Reasoning for the next test: 1. Let's test if the rule involves a simple arithmetic relationship. 2. We'll use numbers where the third is the sum of the first two. 3. We'll use small integers for clarity. Test Case: (2, 3, 5) & (2.0, 3.0, 5.0): False. \\
\cmidrule(r){1-3}
7-8 & ... & ... \\
\cmidrule(r){1-3}
9 & We're still not getting a True result. Let's try a different approach and consider some number theory concepts. Reasoning for the next test: 1. Let's test if the rule involves prime numbers. 2. We'll use three consecutive prime numbers. 3. This will also test if there's a special relationship with consecutive numbers that are not simply ascending. Test Case: (2, 3, 5) & (2.0, 3.0, 5.0): False. \\
\cmidrule(r){1-3}
10 & We've already tested (2, 3, 5) before, and it's still False. Let's try a different number theory concept... & ... \\
\cmidrule(r){1-3}
\end{tabular}
\end{table}

\clearpage

\section{Full Conversation Examples}
\label{appendix:full-convos}

\subsection{Doom Loop on Coding Task}
\label{appendix:doom-loop-example}

Below we provide a full example of a ``doom loop'' interacting with an LLM in a multi-turn conversation in a real world coding task. We use llama 3 8b to generate code for a pytorch variational autoencoder, and after 2 turns of debugging it repeats an already mentioned suggestion.

\begin{quote}

\textbf{\#\#\# User:}

Can you write me some code which trains a VAE on cifar10 using pytorch? \\

\textbf{\#\#\# Llama3-8b-8192:}

Here's a basic example of how you can train a Variational Autoencoder (VAE) on CIFAR-10 using PyTorch:

\begin{Verbatim}[breaklines=true]
```Python
import torch
import torch.nn as nn
import torch.optim as optim
from torchvision import datasets, transforms
from torch.utils.data import DataLoader
import numpy as np

# Define the VAE model
class VAE(nn.Module):
    def __init__(self):
        super(VAE, self).__init__()
        self.encoder = nn.Sequential(
            nn.Conv2d(3, 32, kernel_size=3),
            nn.ReLU(),
            nn.MaxPool2d(2, 2),
            nn.Conv2d(32, 64, kernel_size=3),
            nn.ReLU(),
            nn.MaxPool2d(2, 2),
            nn.Flatten()
        )
        self.fc1 = nn.Linear(64*4*4, 128)
        self.fc2 = nn.Linear(128, 128)
        self.fc3 = nn.Linear(128, 64*4*4)
        self.fc4 = nn.Linear(64*4*4, 3*32*32)

    def reparameterize(self, mu, logvar):
        std = torch.exp(0.5*logvar)
        eps = torch.randn_like(std)
        return mu + eps*std

    def encode(self, x):
        h1 = self.encoder(x)
        h2 = torch.relu(self.fc1(h1))
        h3 = torch.relu(self.fc2(h2))
        mu = self.fc3(h3)
        logvar = self.fc4(h3)
        return mu, logvar

    def decode(self, z):
        h1 = torch.relu(self.fc4(z))
        h2 = torch.relu(self.fc3(h1))
        h3 = torch.relu(self.fc2(h2))
        h4 = torch.relu(self.fc1(h3))
        return torch.sigmoid(self.fc4(h4))

    def forward(self, x):
        mu, logvar = self.encode(x)
        z = self.reparameterize(mu, logvar)
        return self.decode(z), mu, logvar

# Define the loss function
def loss_function(recon_x, x, mu, logvar):
    BCE = nn.functional.binary_cross_entropy(recon_x, x.view(-1, 3*32*32), reduction='sum')
    KLD = 0.5 * torch.sum(1 + logvar - mu.pow(2) - logvar.exp())
    return BCE + KLD

# Load the CIFAR-10 dataset
transform = transforms.Compose([transforms.ToTensor()])
trainset = datasets.CIFAR10(root='./data', train=True, download=True, transform=transform)
trainloader = DataLoader(trainset, batch_size=64, shuffle=True)

testset = datasets.CIFAR10(root='./data', train=False, download=True, transform=transform)
testloader = DataLoader(testset, batch_size=64, shuffle=False)

# Initialize the VAE model and the optimizer
vae = VAE()
optimizer = optim.Adam(vae.parameters(), lr=0.001)

# Train the VAE
for epoch in range(100):
    for i, (x, _) in enumerate(trainloader):
        x = x.to(device)
        optimizer.zero_grad()
        recon_x, mu, logvar = vae(x)
        loss = loss_function(recon_x, x, mu, logvar)
        loss.backward()
        optimizer.step()
        if i % 100 == 0:
            print('Epoch: %d, Batch: %d, Loss: %.3f' % (epoch+1, i+1, loss.item()))

    # Test the VAE
    test_loss = 0
    with torch.no_grad():
        for x, _ in testloader:
            x = x.to(device)
            recon_x, mu, logvar = vae(x)
            loss = loss_function(recon_x, x, mu, logvar)
            test_loss += loss.item()
    test_loss /= len(testloader)
    print('Test Loss: %.3f' % test_loss)
```
\end{Verbatim} 

This code defines a VAE model, loads the CIFAR-10 dataset, and trains the model using the Adam optimizer. The loss function is the sum of the binary cross-entropy loss and the Kullback-Leibler divergence. The model is trained for 100 epochs, and the test loss is printed after each epoch.

Note that you need to install the `torchvision` package to use the CIFAR-10 dataset. You can install it using `pip install torchvision`. Also, you need to specify the device (GPU or CPU) where you want to run the model. You can do this by adding \Verb[breaklines]|`device = torch.device('cuda' if torch.cuda.is_available() else 'cpu')`| before the training loop. \\

\textbf{\#\#\# User:}

I'm getting the following error trying to run this code:

\begin{Verbatim}[breaklines=true]
RuntimeError Traceback (most recent call last)
<ipython-input-3-471d3f454439> in <cell line: 70>()
     72         x = x.to(device)
     73         optimizer.zero_grad()
---> 74         recon_x, mu, logvar = vae(x)
     75         loss = loss_function(recon_x, x, 
                                     mu, logvar)
     76         loss.backward()

6 frames
in forward(self, input)
    115 
    116     def forward(self, input: Tensor) -> Tensor:
--> 117         return F.linear(input, self.weight, self.bias)
    118 
    119     def extra_repr(self) -> str:

RuntimeError: mat1 and mat2 shapes cannot be multiplied (64x2304 and 1024x128)    
\end{Verbatim}

\textbf{\#\#\# llama3-8b-8192:}

The error message indicates that there is a mismatch between the shapes of the input and the weights in one of the fully connected layers in your VAE model. 

The error message specifically says that the shapes are 64x2304 and 1024x128, which means that the input has a shape of (64, 2304) and the weights have a shape of (1024, 128).

The issue is likely due to the fact that the input to the fully connected layer is not what you expect. In the `encode` method of your VAE model, you have the following lines:

\begin{verbatim}
```
h1 = self.encoder(x)
h2 = torch.relu(self.fc1(h1))
h3 = torch.relu(self.fc2(h2))
mu = self.fc3(h3)
logvar = self.fc4(h3)
```
\end{verbatim}

The `h3` tensor has a shape of (64, 128) because `self.fc2` is a fully connected layer with 128 output units. However, `self.fc3` and `self.fc4` are also fully connected layers, and they are expecting an input with a shape of \verb|(batch_size, 128)| where \verb|batch_size| is the number of samples in the batch.

In your case, the batch size is 64, so the input to `self.fc3` and `self.fc4` should be (64, 128), not (64, 128) as it is now.

To fix this issue, you can try to flatten the output of `self.encoder` before passing it to the fully connected layers. You can do this by adding the following line after `h1 = self.encoder(x)`:

\begin{verbatim}
```
h1 = h1.view(-1, 64*4*4)
```
\end{verbatim}

This will flatten the output of `self.encoder` into a tensor with a shape of \verb|(batch_size, 64*4*4)|, which should be compatible with the weights of `self.fc1`, `self.fc2`, `self.fc3`, and `self.fc4`. \\

\textbf{\#\#\# User:}

Okay, I added that line. Now I'm getting a similar error with different values:

\Verb[breaklines]|RuntimeError: mat1 and mat2 shapes cannot be multiplied (144x128 and 1024x3072)| \\

\textbf{\#\#\# llama3-8b-8192:}

The error message indicates that there is a mismatch between the shapes of the input and the weights in one of the fully connected layers in your VAE model. 

The error message specifically says that the shapes are 144x128 and 1024x3072, which means that the input has a shape of (144, 128) and the weights have a shape of (1024, 3072).

The issue is likely due to the fact that the input to the fully connected layer is not what you expect. In the `encode` method of your VAE model, you have the following lines:

\begin{verbatim}
```
h1 = self.encoder(x)
h2 = torch.relu(self.fc1(h1))
h3 = torch.relu(self.fc2(h2))
mu = self.fc3(h3)
logvar = self.fc4(h3)
```
\end{verbatim}

The `h3` tensor has a shape of (64, 128) because `self.fc2` is a fully connected layer with 128 output units. However, `self.fc3` and `self.fc4` are also fully connected layers, and they are expecting an input with a shape of \verb|(batch_size, 128)| where \verb|batch_size| is the number of samples in the batch.

In your case, the batch size is 64, so the input to `self.fc3` and `self.fc4` should be (64, 128), not (64, 128) as it is now.

To fix this issue, you can try to flatten the output of `self.encoder` before passing it to the fully connected layers. You can do this by adding the following line after `h1 = self.encoder(x)`:

\begin{verbatim}
```
h1 = h1.view(-1, 128)
```
\end{verbatim}

This will flatten the output of `self.encoder` into a tensor with a shape of \verb|(batch_size, 128)|, which should be compatible with the weights of `self.fc1`, `self.fc2`, `self.fc3`, and `self.fc4`.

However, the error message also mentions that the input has a shape of 144x128, which is not compatible with the weights of the fully connected layer. This suggests that there may be an issue with the architecture of your VAE model.

To fix this issue, you can try to adjust the architecture of your VAE model to ensure that the input to the fully connected layers is compatible with the weights.

\end{quote}


\end{document}